\documentclass{article}

    \PassOptionsToPackage{numbers, compress}{natbib}

 \usepackage[main, final]{neurips_2026}

\usepackage[utf8]{inputenc} %
\usepackage[T1]{fontenc}    %
\usepackage{hyperref}       %
\usepackage{url}            %
\usepackage{booktabs}       %
\usepackage{amsfonts}       %
\usepackage{nicefrac}       %
\usepackage{microtype}      %
\usepackage{xcolor}         %
\usepackage{amsmath}
\usepackage{amssymb}
\usepackage{mathtools}
\usepackage{amsthm}
\usepackage{multirow}
\usepackage{subcaption}
\usepackage{tablefootnote}
\usepackage{enumitem}
\usepackage{wrapfig}
\usepackage{makecell}
\usepackage[linesnumbered,ruled]{algorithm2e}
\usepackage{kotex}

\SetCommentSty{mycommfont}
\usepackage{subfiles}
\usepackage{titletoc}
\newcommand\DoToC{%
  \startcontents
  \printcontents{}{1}{\vskip3pt\hrule\vskip5pt}
  \vskip15pt\hrule\vskip5pt
}
\usepackage{colortbl}
\definecolor{gainsboro}{RGB}{233,233,233}

\usepackage{xcolor}
\usepackage{pifont}
\usepackage{graphicx}
\newcommand{\proposed}{\textsf{MSP-LLM}}
\newcommand{\cmark}{\textcolor{blue}{\ding{51}}}%
\newcommand{\xmark}{\textcolor{red}{\ding{55}}}%
\usepackage[most]{tcolorbox}
\usepackage{booktabs}
\usepackage{siunitx}
\title{MSP-LLM: A Structured Large Language Model Framework for Inorganic Material Synthesis Planning}

\author{%
  Heewoong Noh\textsuperscript{1}, Gyoung S. Na\textsuperscript{1},
  Namkyeong Lee\textsuperscript{1}, Chanyoung Park\textsuperscript{1}\thanks{Corresponding author.}\\
  {\normalfont\textsuperscript{1}KAIST}\\
  {\normalfont\ttfamily\{heewoongnoh, ngs0726, namkyeong96, cy.park\}@kaist.ac.kr}
}

\begin{document}

\maketitle

\begin{abstract}
Material synthesis planning (MSP) remains a fundamental and underexplored bottleneck in AI-driven materials discovery, as it requires not only identifying suitable precursor materials but also designing coherent sequences of synthesis operations to realize a target material.  Although several AI-based approaches have been proposed to address isolated subtasks of MSP, a methodology for solving the entire MSP task has yet to be established. We propose \proposed, a structured LLM-based framework that formulates MSP as a two-stage process composed of two constituent subproblems: precursor prediction (PP) and synthesis operation prediction (SOP). Our approach introduces a discrete material class as an intermediate decision variable that organizes both tasks into a chemically consistent decision chain. For SOP, we further incorporate hierarchical precursor types as synthesis-relevant inductive biases and employ precursor constraint factorization that preserves precursor-related information in the autoregressive decoding state. Extensive experiments show that \proposed~consistently outperforms existing methods on both PP and SOP, as well as on the MSP task, demonstrating an effective and scalable framework with practical potential for autonomous MSP. The source code for \proposed~is available at~\textcolor{magenta}{\url{https://github.com/HeewoongNoh/MSP-LLM}}.

\end{abstract}

\section{Introduction}
\label{sec:introduction}
Recently, Artificial Intelligence (AI) has achieved notable successes in condensed matter physics and materials science, primarily by enabling a high-throughput screening at the scales of real-world materials spaces \cite{ai_sci_discov1,ai_sci_discov2,ml_mat_ht1, lee2023density}.
After exploring the materials space, we are then confronted with the next scientific challenge of designing a synthesis plan to realize the discovered materials \cite{mat_sci_auto_lab}. However, most existing machine learning methods have primarily focused on exploring the materials space while overlooking the next challenge in \textbf{material synthesis planning (MSP)} \cite{msp,computer_assisted_msp}, and this oversight remains a bottleneck of AI for science in real-world materials discovery. Although numerous synthesis planning methods have been developed for organic compounds, benefiting from well-specified molecular structures and well-defined reaction rules, MSP remains a long-standing challenge in AI for science due to two reasons: (1) The lack of atomic structures for the product materials causes multiple synthesis paths on polymorphic structures, thereby substantially increasing uncertainty in MSP \cite{na2023artificial,llm_msp1}. 
Moreover, (2) fully specifying the synthesis paths based solely on a target material's composition remains a challenge, as compositional data alone fails to provide a comprehensive blueprint for the necessary chemical reactions \cite{noh2024retrieval,prein2025language}.

In real-world applications, MSP is explored within a materials space defined at the level of chemical compositions \cite{na2023artificial,llm_msp1,llm_msp2}, and requires predicting a set of precursor materials $P$ and a sequence of synthesis operations $O$ for a given desired product material. Here, $P \in 2^{\mathcal{M}} \setminus\{\emptyset\}$ denotes a set of precursor materials, where $\mathcal{M}$ denotes the set of materials, and $O \in \mathcal{O}^*$ denotes a finite-length synthesis operation sequence. Formally, the MSP task can be defined as $f : \mathcal{M} \rightarrow 2^{\mathcal{M}} \times \mathcal{O}^*$,
which maps a given product material to a set of precursor materials and a corresponding synthesis operation sequence.
However, due to the inherent challenges of MSP, most existing studies have addressed only partial aspects of MSP, rather than providing a framework that solves MSP in its entirety. In particular, prior works typically focus on one of the following two subproblems: 
(1) \textbf{precursor prediction (PP)}, which predicts a set of precursor materials from a given product material,
$f_p : \mathcal{M} \rightarrow 2^{\mathcal{M}}$,
or (2) \textbf{synthesis operation prediction (SOP)}, which aims to predict a sequence of synthesis operations given a product material and a set of precursor materials,
$f_o : \mathcal{M} \times 2^{\mathcal{M}} \rightarrow \mathcal{O}^*$.

Among these subproblems, SOP remains largely underexplored and has been studied only in narrowly scoped settings, typically restricted to specific material systems rather than general MSP \cite{na2023artificial, pan2025diffsyn}. 
By contrast, most existing efforts have focused on PP, yet these approaches often rely on complex retrieval pipelines \cite{he2023precursor,noh2024retrieval} or access to proprietary LLM APIs \cite{kim2024large, prein2025language}, which limits their scalability and reproducibility.
These limitations underscore the need for the scalable modeling framework that can address both SOP and PP within a general MSP setting.

Recently, large language models (LLMs) fine-tuned for domain-specific scientific tasks have demonstrated strong capabilities across scientific domains, particularly in materials science, including generation of candidate materials \cite{gruver2024fine} and emulation of scientific discovery processes \cite{ding2024matexpert}. This modeling framework is particularly well-suited to MSP, where all relevant input information is naturally represented as text in the form of chemical compositions \cite{prein2025language}. By leveraging LLMs that have been pre-trained on large-scale and diverse textual corpora, LLM-based MSP models can potentially access and utilize extensive background knowledge, enabling more informed and physically grounded decisions \cite{bran2025chemical, rueda2025understanding} in both precursor selection and synthesis operation planning.

However, despite their promise, effective strategies for fine-tuning LLMs to address MSP remain largely unexplored. Specifically, no existing work has established a principled fine-tuning strategy that enables LLMs to perform PP and SOP in a structured MSP framework.

In this paper, we propose a structured LLM-based framework, called \proposed, for solving complex MSP tasks. We develop novel task-specific LLM fine-tuning strategies for the two constituent subproblems of MSP, i.e., PP and SOP. Specifically, (i) we introduce a discrete material class variable that captures synthesis-relevant characteristics of materials and employ it in both PP and SOP as an intermediate decision variable. By incorporating this variable, each subtask is organized into a structured decision chain, in which the model conditions its task-specific generation on the material class variable, leading to more accurate  predictions. Furthermore, for SOP, we introduce an explicit conditioning strategy that integrates hierarchical precursor types as synthesis-relevant inductive biases and enforces their effective utilization through  precursor constraint factorization, promoting precursor-conditioned operation generation and more reliable retention of precursor constraints in the decoder state. By integrating LLMs fine-tuned for each subproblem within a single coherent framework, \proposed~performs MSP given target material compositions. Extensive experiments demonstrate that \proposed~significantly outperforms existing methods on both individual subproblems and the MSP task, highlighting its effectiveness and practical applicability to real-world materials discovery.

To the best of our knowledge, \proposed~is the first work to formulate and tackle both PP and SOP within a single MSP framework, and to show that this can be effectively addressed by fine-tuning LLMs in a structured manner.

In this study, we make the following contributions:
\begin{itemize}[leftmargin=.1in, itemsep=1pt, parsep=1pt, topsep=0pt]

    \item We propose \proposed~, a structured framework for general Material Synthesis Planning (MSP) that formulates the task as a two-stage process consisting of (1) precursor prediction (PP) and (2) synthesis operation prediction (SOP), each implemented using fine-tuned LLMs tailored to the respective subtask.
    \looseness=-1
    \item  By incorporating domain-specific inductive biases via a material group decision chain in both PP and SOP, together with an explicit conditioning strategy, our framework effectively leverages precursor information in SOP. We further analyze the mechanisms underlying the resulting performance gains.
    \item Through extensive experiments, we demonstrate that the proposed framework significantly outperforms all baselines on the MSP task as well as its subtasks, highlighting the superiority of our approach.

\end{itemize}

\section{Preliminaries}
\label{sec:prelim}
\noindent{\textbf{Material Synthesis Planning (MSP).}} As introduced in Section~\ref{sec:introduction}, MSP associates a target material with a precursor set and a synthesis operation sequence.
We adopt a two-stage formulation that emulates the experimental workflow of synthesis chemists who first identify feasible precursors and then design operations conditioned on them. While a one-stage approach can, in principle, learn to output both precursors and operations simultaneously, such predictions collapse material selection and process design into a single, opaque step that is difficult to verify, adapt, or safely execute in practice \cite{tang2025risks}. By explicitly separating these decisions, the two-stage formulation better aligns with real-world material synthesis workflows, enabling more verifiable, and practically deployable planning.

\noindent{\textbf{Precursor Prediction (PP).}}
This task aims to identify a set of starting materials required to synthesize a given target material.
In this task, only the chemical composition of the target material $m$ is provided, without any explicit structural information,
and the goal is to predict an appropriate precursor set $P$.
For example, when the target material is $\mathrm{La_{1.2}Ca_{1.8}Mn_{2}O_7}$, the task is to predict a precursor set $\{\mathrm{CaCO_3}, \mathrm{La_2O_3}, \mathrm{MnO_2}\}$. Details of the preprocessing procedure for precursors are provided in Appendix \ref{app: Dataset}.

\noindent{\textbf{Synthesis Operation Prediction (SOP).}}
This task focuses on determining the sequence of processing steps required to synthesize the target material,
given both the target material and its corresponding precursor set.
We define the synthesis operation as  the sequence
$O = (o_1, o_2, \ldots, o_n)$, where each $o_t$ is selected from a predefined set of
seven canonical solid-state synthesis operations:
\emph{mixing}, \emph{drying}, \emph{heating}, \emph{sintering}, \emph{annealing}, \emph{quenching}, and \emph{shaping}.
A valid synthesis route may involve repeating certain operations multiple times,
and the objective is to predict the ordered sequence $O$.
For instance, in the case of $\mathrm{La_{1.2}Ca_{1.8}Mn_{2}O_7}$ with the precursor set
$\{\mathrm{CaCO_3}, \mathrm{La_2O_3}, \mathrm{MnO_2}\}$,
a plausible synthesis sequence is
$\textit{mixing} \rightarrow \textit{heating} \rightarrow \textit{mixing} \rightarrow \textit{shaping} \rightarrow \textit{sintering}$.
Details of the preprocessing of synthesis operations are provided in Appendix \ref{app: Dataset}.

\section{Proposed Method }
\label{sec:methodology}
\begin{figure*}[t]
    \centering
    \includegraphics[width=0.90\linewidth]
    {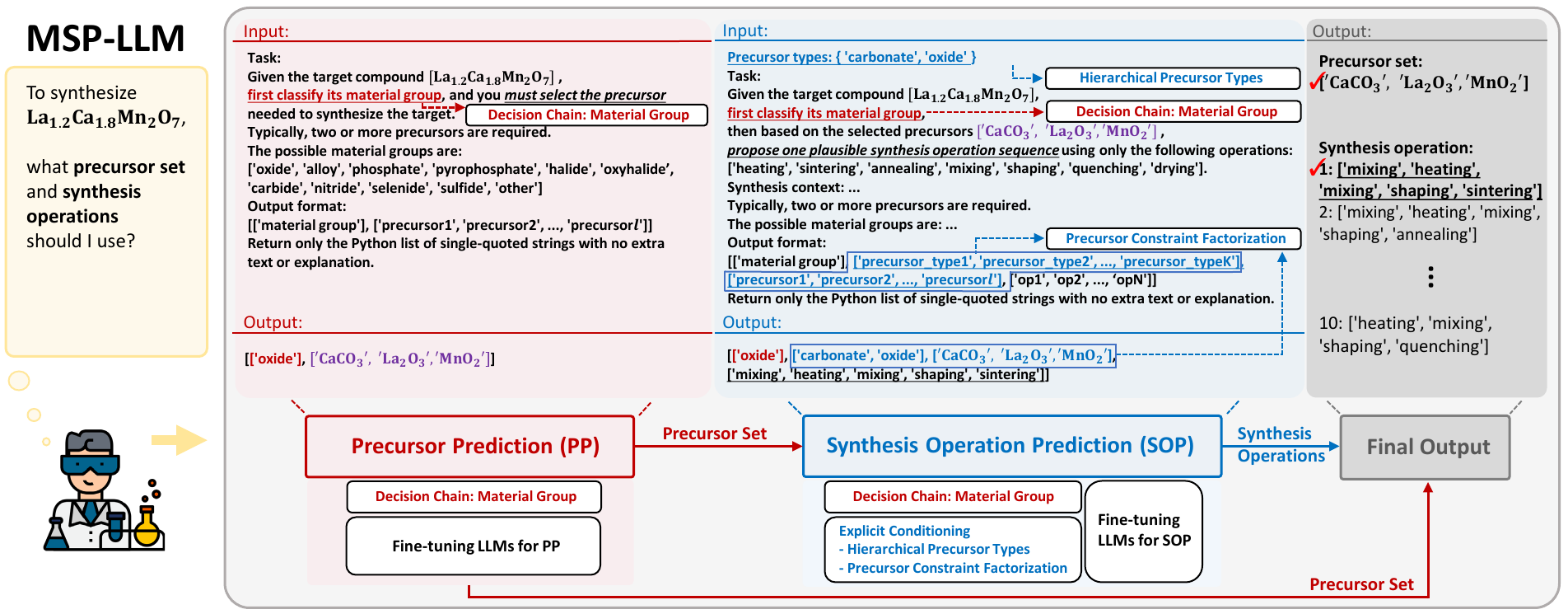}
    \caption{Overall framework of \proposed. Given a target material, \proposed~first predicts the precursor set (PP), and then, conditioned on the predicted precursors, generates the synthesis operation sequence (SOP), resulting in a MSP. The input–output texts above the PP and SOP LLMs illustrate example prompts used during LLM fine-tuning for each task.}
    \label{fig:model architecture}
\end{figure*}

We develop task-specific LLM fine-tuning strategies for the two constituent subproblems of MSP: PP and SOP. To structure these tasks, we introduce a discrete material class variable that induces a decision-chain formulation (Section~\ref{decision chain}). We then describe the PP stage (Section~\ref{PP stage}) and SOP stage (Section~\ref{SOP stage}). We further propose explicit conditioning with hierarchical precursor types, which reinforces precursor-related constraints in the decoder state (Section~\ref{explicit conditioning}), and analyze why this design is effective (Section~\ref{sec:why_pcf}). 
Finally, we describe the MSP inference pipeline (Section~\ref{complete msp}), where the independently trained PP and SOP models are linked sequentially: precursors predicted by PP serve as conditioning inputs to SOP.

\subsection{Decision Chain via Discrete Material Class Variable}
\label{decision chain}

Materials often exhibit synthesis-related characteristics that are correlated with their chemical classes \cite{rao1997new, west2022solid}. While a material group does not uniquely determine a specific synthesis pathway, materials belonging to different groups (e.g., oxides, phosphates, nitrides, sulfides) often exhibit systematic differences in precursor selection, decomposition behavior, and synthesis operations \cite{synth_2}. In practice, domain experts attempt to identify the chemical group of a target material to determine an appropriate synthesis strategy. Inspired by this expert reasoning process, we introduce a decision chain framework that utilizes the material group, denoted as $G$, as a discrete intermediate variable for both Precursor Prediction (PP) and Synthesis Operation Prediction (SOP). In the PP task, given a target material $m$, the model first predicts its material group and subsequently generates its precursor set $P$ conditioned on the predicted material group. Similarly, for SOP, the model predicts the material group $G$ based on the target material $m$ and the predicted precursor set $P$ before generating the synthesis operation sequence $O$. Accordingly, both tasks are formulated as structured decision chains that first infer the material group and then perform task-specific generation. To implement this decision chain, we curate input prompts that explicitly instruct the model to \textbf{"first classify its material group"} before proceeding to the PP/SOP. Consequently, while fine-tuning the LLM, it learns to perform material group classification as a prerequisite step. This approach introduces a task-specific inductive bias, effectively constraining the search space for precursors and operations to a chemically plausible subspace. Note that PP and SOP are implemented using independently fine-tuned LLMs, and are subsequently integrated within a coherent framework where the material group decision chain is applied to both stages.

\subsection{Precursor Prediction (PP) Stage} 
\label{PP stage}
For the PP stage, we design a task-specific prompt to fine-tune the LLM such that the model first classifies the material group $G$ from the target material $m$ and subsequently selects an appropriate precursor set $P$.
Let $X_{\text{PP}}$ denote the complete input prompt, which includes the task description and the chemical formula of the target material $m$. We train the model using a standard language modeling (LM) objective under teacher forcing over the entire output sequence, with the target sequence $\mathbf{y}_{\text{PP}} = (G, P)$. Formally, the training loss is given by $\mathcal{L}_{\text{LM}}^{\text{PP}} = - \sum_{t=1}^{|\mathbf{y}_{\text{PP}}|} \log p_\theta ( y_t \mid X_{\text{PP}}, \mathbf{y}_{\text{PP}<t} )$, where $\mathbf{y}_{\text{PP}<t}$ denotes the ground-truth prefix.

\subsection{Synthesis Operation Prediction (SOP) Stage}
\label{SOP stage}

\begin{wrapfigure}{r}{0.44\textwidth} 
    \centering
    \includegraphics[width=\linewidth]{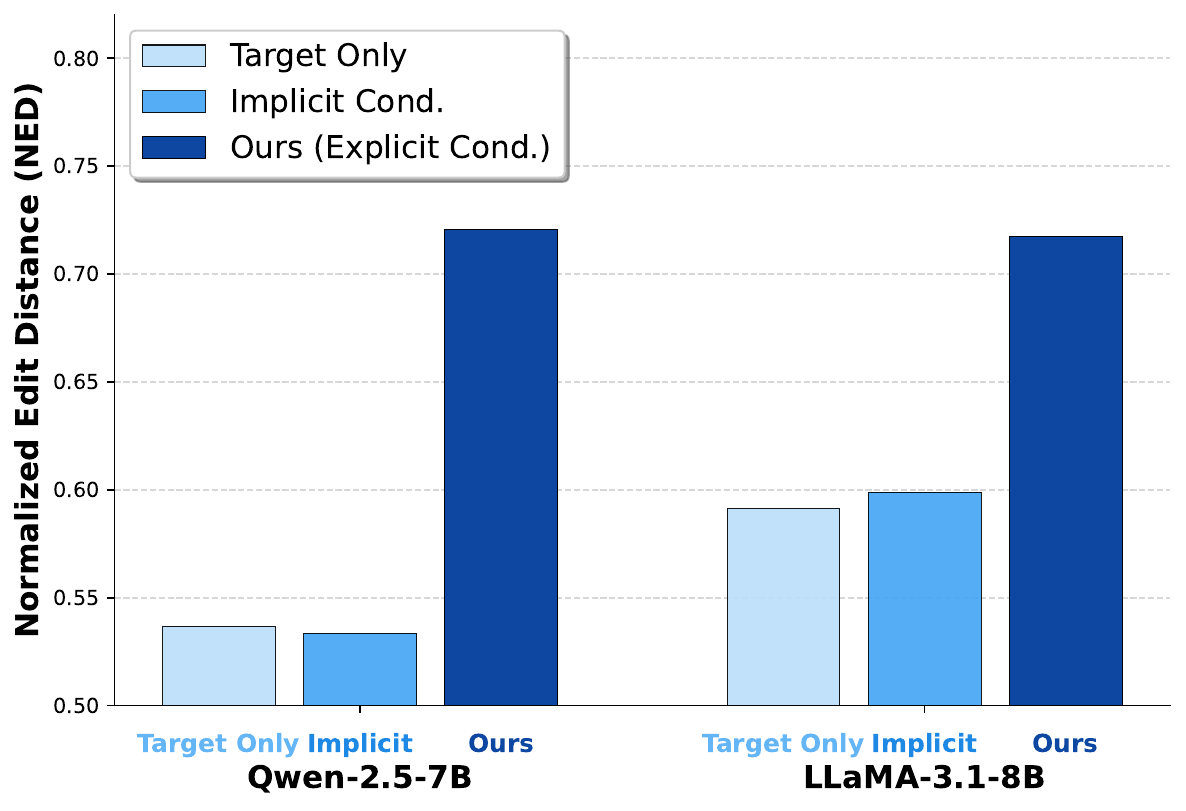}
    \caption{SOP performance under target-only, implicit, and explicit conditioning across two fine-tuned backbone LLMs.}
    \label{fig:motivation}

\end{wrapfigure}
\subsubsection{Challenges of Precursor Utilization}
\label{explicit conditioning}
In SOP, precursors play a pivotal role in determining synthesis operation behavior, as they provide particularly informative cues for materials that require complex, multi-step synthesis procedures~\cite{rao1997new,west2022solid,synth_2}. 
Intuitively, incorporating the precursor set into the input context is therefore expected to improve prediction accuracy compared to conditioning solely on the target material.
To corroborate this, we compare two simple conditioning strategies applied at the LLM input stage: \textbf{target-only} and \textbf{implicit conditioning}, across two backbone LLMs, Qwen-2.5-7B and LLaMA-3.1-8B. 
The \textbf{target-only} approach conditions the LLM on the task description and the target material only, while \textbf{implicit conditioning} additionally incorporates the precursor set alongside the target material information to guide the model. We evaluate SOP performance using Normalized Edit Distance (NED), which measures the token-level similarity between the predicted and ground-truth synthesis operation sequences (note that each synthesis operation is treated as a single token).

However, as illustrated in Figure~\ref{fig:motivation}, contrary to our initial expectations, \textbf{implicit conditioning} does not consistently outperform the \textbf{target-only} baseline, despite the inclusion of additional precursor information.
This observation suggests that merely providing precursor information as a part of the LLM input is insufficient to ensure its effective utilization, especially in complex contexts—such as those involving long sequences or {heterogeneous sequences of chemical formulas, where multiple precursor formulas are listed consecutively.}

Therefore, this motivates the need to explicitly enforce the preservation and utilization of precursor-related information throughout the decoding process. 

To effectively leverage the precursor information, we introduce an \textbf{explicit conditioning} strategy comprising two key components:
(i) extracting hierarchical precursor types from the fine-grained precursor set to capture high-level behavioral constraints (Section~\ref{hierarchical precursor types}), and
(ii) enforcing conditioning via precursor constraint factorization (Section~\ref{precursor constraint factorization}). As shown in Figure~\ref{fig:motivation}, this design leads to substantial and consistent performance gains across both backbone models, highlighting the importance of \textbf{explicit conditioning} over naive implicit input-level conditioning.
Note that, since the PP and SOP modules are trained independently, we use the ground-truth precursor set to condition the SOP module during the model training.

\subsubsection{Hierarchical Precursor Types}
\label{hierarchical precursor types}

In SOP, it is important not only to identify fine-grained operational dependencies, such as the ordering of steps and the need for repeated operations, but also to capture higher-level synthesis trends that govern the overall process. For example, carbonate precursors typically require an initial heating step to remove CO$_2$ prior to phase formation~\cite{synth_1}. This implies that the presence of a carbonate precursor provides a strong behavioral signal indicating the necessity of a heating step in the synthesis route. Building on this insight, we leverage a hierarchical abstraction of the precursor set in the form of precursor types, which capture key synthesis-relevant behaviors such as decomposition, volatility, and reactivity~\cite{synth_2,rao1997new,west2022solid}. Let $P = \{p_1, \ldots, p_l\}$ denote the precursor set for a target material, and $p_i$ denote the $i$-th precursor, where $l$ is the number of precursors in $P$. Each precursor belongs to one of a finite set of precursor types (e.g., carbonate, nitrate, ammonium, phosphate, oxide, or other). We define a simple rule-based mapping function $\tau$ such that $Types = \tau(P)$, where ${Types}$ denotes the set of unique precursor types associated with the precursor set $P$. The resulting ${Types}$ provides a hierarchical, behaviorally meaningful abstraction of $P$ that aligns closely with typical synthesis operation patterns. We denote the comprehensive precursor information as $Z := ({Types}, P)$. For instance, given a precursor set $P=\{\text{CaCO}_3, \text{La}_2\text{O}_3, \text{MnO}_2\}$, the corresponding unique ${Types}$ would be $\{{\text{carbonate}, \text{oxide}}\}$.

\subsubsection{Precursor Constraint Factorization}
\label{precursor constraint factorization}

To ensure the effective utilization of precursor information beyond implicit input-level context, we introduce \emph{precursor constraint factorization (PCF)}, which enforces the model to utilize the precursor information. Unlike naive implicit conditioning, which offers limited training pressure for precursor-specific representations in the decoder state, PCF introduces an explicit constraint that facilitates more consistent utilization of precursor information during synthesis operation generation. Concretely, SOP is formulated as learning a joint conditional distribution over $(G,Z,O)$ given $X_{\text{SOP}}$, where $X_{\text{SOP}} = (\text{task prompt}, m,  Z_{\text{in}})$ and $Z_{\text{in}}$ denotes the ground-truth precursor information provided as part of the input. This induces the following factorization:
\begin{equation}
\small
\begin{aligned}
p_{\theta}(G, Z, O \mid X_{\text{SOP}}) 
&= p_{\theta}(G \mid X_{\text{SOP}})\,
   p_{\theta}(Z \mid X_{\text{SOP}}, G) \\
&\quad  \cdot p_{\theta}(O \mid X_{\text{SOP}}, G, Z).
\end{aligned}
\label{eq:sop_factorization}
\end{equation}
\looseness=-1
This formulation provides a principled decomposition that first enables the model, through the term $p_{\theta}(G \mid X_{\text{SOP}})$, to identify the material group as an intermediate variable corresponding to the decision-chain formulation (Section \ref{decision chain}). The subsequent term $p_{\theta}(Z \mid X_{\text{SOP}}, G)$ explicitly constrains the model to predict and utilize precursor information. Finally, the term $p_{\theta}(O \mid X_{\text{SOP}}, G, Z)$ trains the model to generate the synthesis operation sequence conditioned on both the inferred material group and the predicted precursor variables. We implement this factorization by structuring the output sequence as $\mathbf{y}_{\text{SOP}} = (G, Z, O)$ and optimizing a standard autoregressive language modeling objective,
\begin{equation}
\small
\mathcal{L}_{\text{LM}}^{\text{SOP}} = - \sum_{t=1}^{|\mathbf{y}_{\text{SOP}}|}\log p_{\theta}(y_t \mid X_{\text{SOP}}, \mathbf{y}_{\text{SOP}<t}).
\label{eq:sop_lm}
\end{equation}

During training, the ground truth precursor information $Z_{\text{in}}$ is provided as part of the input context while simultaneously being modeled as an explicit random variable $Z$ in the output sequence. This requires the decoder state to remain sufficient for predicting $Z$ even when $Z$ is already observed in the input.
Under this formulation, minimizing the SOP loss jointly constrains the model to accurately predict both $Z$ and $O$, thereby discouraging solutions in which precursor-specific information is not retained in the decoder representation during operation generation. 
Consequently, during operation generation, the decoder state must remain aligned with precursor-specific constraints, enforcing retention and utilization of $Z$ rather than treating precursor information as a passive component of the input context.

\subsection{MSP Inference Pipeline}
\label{complete msp}
At inference time, \proposed~ produces a synthesis plan from the target material composition $m$. Given this target, (1) the fine-tuned PP model first generates a ranked set of candidate precursor combinations. (2) Conditioned on a selected precursor set, the fine-tuned SOP model then produces coherent sequences of synthesis operations that specify how to realize the target material using the predicted precursors. By integrating these steps, the framework effectively achieves the MSP task.

\subsection{Why Precursor Constraint Factorization Works}
\label{sec:why_pcf}

\paragraph{Setup.} For clarity, we omit $G$ and focus on the precursor information $Z$ and the operation sequence $O$. We decompose the input context as $X_{\text{SOP}} = (X, Z_{\text{in}})$, where $X = (\text{task prompt}, m)$ and $Z_{\text{in}}$ is the ground-truth precursor information provided in the prompt. We distinguish $Z_{\text{in}}$ from the structured precursor variable $Z := (\text{Types}, P)$, which PCF treats as an explicit random variable in the output sequence. Implicit conditioning optimizes $-\log p_\theta(O \mid X_{\text{SOP}})$, whereas PCF optimizes $-\log p_\theta(Z, O \mid X_{\text{SOP}})$.

\paragraph{Marginalization under implicit conditioning.}
Since implicit conditioning does not require the model to distinguish between different precursor-specific generation mechanisms (i.e., $Z_{in}$ and $Z$), it admits solutions that achieve low training loss while approximating the marginalized distribution $p(O \mid X) = \sum_{z} p(O \mid X, Z=z)\,p(Z=z \mid X)$, in which distinct precursor-conditioned patterns collapse into a single aggregated distribution and the decoder state is not required to retain instance-specific precursor information.
\paragraph{Precursor retention under PCF.}
\looseness=-1
The distinction between the marginalized distribution $p(O \mid X)$ and the precursor-conditioned distribution $p(O \mid X, Z)$ can be formalized using a basic property of conditional entropy: $H(O \mid X, Z) \leq H(O \mid X)$, with equality if and only if $O \perp Z \mid X$. When precursor information is informative for operations, i.e., $I(O; Z \mid X) > 0$, modeling $p(O \mid X, Z)$ admits strictly lower uncertainty than the marginalized predictor. Realizing this benefit, however, requires the decoder state to retain precursor information during operation generation. PCF provides this by training the model to predict $Z$ as part of the output sequence, acting as an auxiliary objective that forces the decoder to retain precursor-specific features. This mitigates collapse into the marginalized predictor and narrows the gap between the learned model and $p(O \mid X, Z)$, yielding operation sequences that better reflect the given precursors.

\paragraph{Empirical verification.} 
\begin{wraptable}{r}{0.42\linewidth}
\centering
\caption{Model probing accuracy for precursor types (Unseen formula, LLaMA-3.1-8B). $\Delta$ denotes improvement over the previous row.}
\label{tab:probing}
\small
\setlength{\tabcolsep}{4pt}
\renewcommand{\arraystretch}{1.05}
\begin{tabular*}{\linewidth}{@{\extracolsep{\fill}}lcc@{}}
\toprule
\textbf{Method} & \textbf{Acc.} & \textbf{$\Delta$} \\
\midrule
Target only      & 0.6502 & --- \\
Implicit         & 0.6838 & \textit{+3.4\,pp} \\
Explicit (PCF)   & \textbf{0.8479} & \textbf{+16.4\,pp} \\
\bottomrule
\end{tabular*}
\end{wraptable}

We further verify this mechanism by probing the decoder hidden states directly. We compare three settings: target-only (no precursors), implicit conditioning, and explicit conditioning with PCF; the latter two exclude hierarchical precursor types to isolate the effect of PCF. We extract mean-pooled hidden states during $O$ generation and train a linear classifier for multi-label precursor type prediction (6 classes). As shown in Table~\ref{tab:probing}, implicit conditioning improves over the target-only baseline by only $3.4$ percentage points despite receiving precursor information as input, indicating that input-side precursor information is largely marginalized in the decoder state. In contrast, PCF yields a substantial $16.4$ percentage-point gain over implicit conditioning, confirming that explicit prediction of precursor variables actively preserves precursor-specific information in the decoder state. Together with the consistent SOP gains in Figure~\ref{fig:motivation}, this provides direct evidence that PCF achieves the precursor-aware decoding behavior predicted by our analysis. A counterfactual precursor-substitution analysis (Appendix~\ref{app:counterfactual}) further shows that PCF makes the generated operations respond to precursor changes in a chemically consistent direction, whereas implicit conditioning largely does not.

\section{Experiments}
\label{sec:experiments}
\looseness=-1
We evaluate the effectiveness of \proposed~ across three settings: the PP task (Section \ref{ex_pp}), which predicts precursor sets from a target material; the SOP task (Section \ref{ex_op}), which predicts synthesis operations given the target material and ground-truth precursors; and the MSP task (Section \ref{ex_msp}), which predicts both precursors and synthesis operations simultaneously from the target material composition.

\noindent{\textbf{Datasets.}} 
We utilize an inorganic material synthesis dataset that extracts synthesis recipes from materials science papers \cite{kononova2019text}, which has been used in prior studies focusing on PP \cite{noh2024retrieval, kim2024large}. To address the MSP task, we curate a subset of the data that contains full target, precursor, and synthesis operation information, resulting in 10,851 real-world synthesis records. For PP, we restrict the precursors to those with a frequency of at least five, yielding a set of 288 precursors.
For synthesis operations, the original scheme consists of five categories, namely \emph{mixing}, \emph{quenching}, \emph{heating}, \emph{drying}, and \emph{shaping}. In the SOP setting, we refine this scheme by decomposing \emph{heating}, which conflates procedurally distinct operations, into \emph{sintering}, \emph{annealing}, and a general \emph{heating} category, yielding seven operation types that better reflect real-world synthesis procedures. Additional details about the dataset are provided in Appendix \ref{app: Dataset}.

\noindent{\textbf{Implementation Details.}} We construct \proposed~by fine-tuning LLMs to perform both PP and SOP. We leverage open-source models LLaMA-3.1-8B \cite{grattafiori2024llama} and Qwen 2.5-7B \cite{qwen2025qwen25technicalreport}, and apply QLoRA \cite{dettmers2023qlora} to enable efficient fine-tuning with limited computational resources. 
As detailed in Section \ref{complete msp}, we fine-tune and evaluate the LLMs independently for each task. For the MSP task, these separately fine-tuned models are deployed simultaneously to generate the final output.
Additional implementation details are provided in Appendix \ref{app: Implementation}.

\noindent{\textbf{Evaluation Protocol.}} We consider two dataset split strategies with an 8:1:1 train/validation/test ratio for experiments. Each sample is associated with a unique precursor set and synthesis operation sequence. The first is a random split, in which target formulas may appear across different splits, whereas each occurrence corresponds to a different precursor set and synthesis operation sequence due to the one-to-many mapping. The second is a target-disjoint split, where target formulas are strictly separated across different splits, enabling evaluation on previously unseen target materials. Additional details are provided in Appendix \ref{app: Dataset}. For SOP, all methods additionally receive synthesis context extracted from the titles of source papers (Appendix~\ref{app:preprocessing}); results without this context are reported in Appendix~\ref{app:synthesis_context}. 

\subsection{Precursor Prediction (PP) Task}
\label{ex_pp}

\noindent{\textbf{Baseline Methods.}} We prepare two types of baselines:
(1) Specialized models for PP: we consider retrieval-based approaches that leverage reference material information through retrieval mechanisms, including the method of \citet{he2023precursor}, which uses a composition-vector–based retrieval scheme, and Retrieval-Retro \cite{noh2024retrieval}, a state-of-the-art model that combines advanced retrieval mechanisms with attention.
(2) LLM baselines: we follow the prompt design of \citet{prein2025language} to evaluate proprietary LLMs on PP, and conduct experiments with GPT-4o-mini, GPT-4o, GPT-5.1, and Claude Haiku 4.5. Here, we report the performance of \proposed~using an LLM that is independently trained specifically for the PP task.

\noindent{\textbf{Evaluation Protocol.}} Following prior work~\cite{prein2025language, noh2024retrieval}, we adopt the Top-$K$ Exact Match metric to evaluate precursor prediction performance. 

\noindent{\textbf{Empirical Results.}} In Table~\ref{tab:main_precursor}, we observe that both specialized models and API-based LLMs exhibit limited capability in precursor prediction. 
Specifically, while the retrieval-based specialized model (Retrieval-Retro) achieves performance comparable to GPT-4o, it relies on an external retrieval system and lacks flexibility in adapting the size of the predicted precursor set, which constrains its overall accuracy.
Moreover, while increasing model scale and the number of in-context examples improves API-based LLMs, a substantial performance gap remains.
In contrast, our framework, regardless of the underlying LLM backbone, consistently outperforms all baselines across all Top-$K$ levels. 
Since these LLMs are evaluated with few-shot prompting, we further compare against GPT-4o-mini fine-tuned on the same training data, over which \proposed~still achieves higher performance on PP, SOP, and MSP (Appendix~\ref{app:extended_api}).
These results demonstrate that structured, task-specific fine-tuning and domain-aware inductive biases enable more effective learning of precursor selection constraints than either prompt-based conditioning or retrieval-dependent pipelines.

\begin{table*}[t]
\centering
\small

\begingroup
\setlength{\aboverulesep}{0pt}
\setlength{\belowrulesep}{0pt}
\renewcommand{\arraystretch}{1.3}

\begin{minipage}[t]{0.49\textwidth}
    \centering
    \caption{PP performance on two dataset splits.}
    \label{tab:main_precursor}

    \resizebox{0.96\linewidth}{!}{%
    \begin{tabular}{l cccc c cccc}
    \toprule
    \textbf{Method} & \multicolumn{4}{c}{\textbf{Split 1}} & & \multicolumn{4}{c}{\textbf{Split 2}} \\
    \cmidrule(lr){2-5} \cmidrule(lr){7-10}
    & \textbf{Top-1} & \textbf{Top-3} & \textbf{Top-5} & \textbf{Top-10} & & \textbf{Top-1} & \textbf{Top-3} & \textbf{Top-5} & \textbf{Top-10} \\
    \midrule
    
    \rowcolor{gray!12}
    \multicolumn{5}{l}{\textit{Specialized Models}} & & \multicolumn{4}{l}{} \\
    \midrule
    \citet{he2023precursor} & 53.11 & 60.05 & 61.06 & 63.07 & & 54.01 & 61.10 & 62.69 & 65.02 \\
    Retrieval-Retro~\cite{noh2024retrieval} & 60.88 & 69.38 & 70.93 & 74.13 & & 66.32 & 74.16 & 75.84 & 78.54 \\
    
    \midrule
    \rowcolor{gray!12}
    \multicolumn{5}{l}{\textit{Large Language Models (API, few-shot prompting)}} & & \multicolumn{4}{l}{} \\
    \midrule
    GPT-4o-mini (20-shot) & 56.58 & 65.36 & 67.82 & 70.66 & & 54.85 & 63.40 & 67.91 & 70.34 \\
    GPT-4o-mini (40-shot) & 55.21 & 63.44 & 66.27 & 69.10 & & 55.97 & 64.27 & 67.82 & 71.36 \\
    GPT-4o (20-shot) & 60.05 & 68.56 & 71.76 & 75.05 & & 60.17 & 69.31 & 73.04 & 76.31 \\
    GPT-4o (40-shot) & 61.79 & 70.48 & 73.40 & 76.32 & & 61.57 & 71.27 & 73.97 & 77.05 \\
    GPT-5.1 (20-shot) & 64.72 & 77.42 & 80.53 & 83.27 & & 67.07 & 77.61 & 82.37 & 85.17 \\
    GPT-5.1 (40-shot) & 64.08 & 76.33 & 79.62 & 83.46 & & 66.60 & 77.89 & 82.74 & 86.01 \\
    Claude Haiku 4.5 (20-shot) & 55.03 & 68.37 & 74.50 & 78.88 & & 56.44 & 68.84 & 73.13 & 78.92 \\
    Claude Haiku 4.5 (40-shot) & 57.31 & 68.74 & 74.31 & 78.43 & & 56.53 & 71.64 & 76.12 & 81.81 \\
    \midrule\midrule
    \textbf{\proposed~(Qwen-2.5-7B)} & \textbf{71.02} & \underline{85.37} & \underline{90.49} & \underline{94.06} & & \underline{71.92} & \underline{88.62} & \underline{92.72} & \underline{95.53} \\
    \textbf{\proposed~(LLaMA-3.1-8B)} & \underline{70.75} & \textbf{85.56} & \textbf{91.68} & \textbf{94.97} & & \textbf{72.85} & \textbf{88.99} & \textbf{93.56} & \textbf{96.36} \\
    \bottomrule
    \end{tabular}
    }
\end{minipage}
\hfill %
\begin{minipage}[t]{0.49\textwidth}
    \centering
    \caption{SOP performance on two dataset splits.}
    \label{tab:main_operation}
    
    \resizebox{1.0\linewidth}{!}{%
    \begin{tabular}{l cccc c cccc}
    \toprule
    \textbf{Method} & \multicolumn{4}{c}{\textbf{Split 1}} & & \multicolumn{4}{c}{\textbf{Split 2}} \\
    \cmidrule(lr){2-5} \cmidrule(lr){7-10}
    & \textbf{Top-10} & \textbf{NED}$\uparrow$ & \textbf{LCS}$\uparrow$ & \textbf{F-1}$\uparrow$ & & \textbf{Top-10} & \textbf{NED}$\uparrow$ & \textbf{LCS}$\uparrow$ & \textbf{F-1}$\uparrow$ \\
    \midrule
    
    \rowcolor{gray!12}
    \multicolumn{5}{l}{\textit{Specialized Models}} & & \multicolumn{4}{l}{} \\
    \midrule
    Retrieval based (\citet{he2023precursor})& 10.4 & 0.6072 & 0.7194 & 0.7596 & & 11.19 & 0.6030 & 0.7198 & 0.7564 \\
    SPENDE \cite{na2023artificial} & N/A & 0.4508 & 0.5824 & 0.6013 & & N/A & 0.4522 & 0.5821 & 0.6010 \\
    
    \midrule
    \rowcolor{gray!12}
    \multicolumn{5}{l}{\textit{Large Language Models (API, few-shot prompting)}} & & \multicolumn{4}{l}{} \\
    \midrule
    GPT-4o-mini (20-shot) & 5.85 & 0.5717 & 0.6992 & 0.7355 & & 7.09 & 0.5847 & 0.7089 & 0.7407 \\
    GPT-4o-mini (40-shot) & 5.94 & 0.5844 & 0.7092 & 0.7454 & & 7.93 & 0.5934 & 0.7141 & 0.7442 \\
    GPT-4o (20-shot) & 5.03 & 0.5654 & 0.6925 & 0.7219 & & 3.82 & 0.5670 & 0.6865 & 0.7166 \\
    GPT-4o (40-shot) & 4.34 & 0.5693 & 0.6890 & 0.7195 & & 4.94 & 0.5749 & 0.6924 & 0.7245 \\
    GPT-5.1 (20-shot) & 2.93 & 0.5334 & 0.6583 & 0.7101 & & 2.33 & 0.5301 & 0.6547 & 0.6968 \\
    GPT-5.1 (40-shot) & 2.74 & 0.5434 & 0.6660 & 0.7146 & & 2.33 & 0.5261 & 0.6498 & 0.6925 \\
    Claude Haiku 4.5 (20-shot) & 2.56 & 0.5534 & 0.6727 & 0.7174 & & 1.68 & 0.5439 & 0.6656 & 0.7017 \\
    Claude Haiku 4.5 (40-shot) & 4.11 & 0.5669 & 0.6801 & 0.7172 & & 3.26 & 0.5569 & 0.6730 & 0.7037 \\
    
    \midrule\midrule
    \textbf{\proposed~(Qwen-2.5-7B)} & \textbf{26.51} & \textbf{0.7207} & \textbf{0.8100} & \textbf{0.8359} & & \textbf{32.18} & \underline{0.7420} & \textbf{0.8240} & \underline{0.8493} \\
    \textbf{\proposed~(LLaMA-3.1-8B)} & \textbf{26.51} & \underline{0.7174} & \underline{0.8075} & \underline{0.8353} & & \underline{31.44} & \textbf{0.7424} & \underline{0.8235} & \textbf{0.8498} \\
    \bottomrule
    \end{tabular}
    }
\end{minipage}

\endgroup
\end{table*}

\subsection{Synthesis Operation Prediction (SOP) Task}
\label{ex_op}

\noindent{\textbf{Baseline Methods.}} We evaluate two categories of baselines: (1) Specialized models for SOP: to emulate real experimental practices of chemists, who search for materials similar to a target and adapt their reported synthesis procedures, we use the retriever proposed by \citet{he2023precursor} to identify similar materials in the training set and take their associated synthesis operations as predictions. In addition, we evaluate SPENDE \cite{na2023artificial} as a specialized baseline. 
(2) LLM baselines: we design task-specific prompts to evaluate proprietary LLMs on SOP, and conduct experiments with GPT-4o-mini, GPT-4o, GPT-5.1, and Claude Haiku 4.5. 

\looseness=-1
\noindent{\textbf{Evaluation Protocol.}} Since SOP is a sequence generation task, we adopt the Top-$K$ Exact Match metric as a strict evaluation criterion that requires both the operation set and the exact ordering of the sequence to match the ground truth. To provide a more fine-grained assessment, we additionally report the Normalized Edit Distance (NED), which measures the similarity between predicted and ground-truth operation sequences, and the Longest Common Subsequence (LCS), which captures the length of the longest ordered subsequence shared with the ground truth. Finally, to evaluate content-level coverage independent of ordering, we use the multiset F-1 score (F-1), which measures the overlap between the predicted and reference operation elements while accounting for duplicates. For these sequence-level metrics, we report the maximum value among the Top-10 candidate predictions. For SPENDE~\cite{na2023artificial}, which produces only a single output per input, top-10 metrics are not applicable; instead, we report alternative evaluation metrics computed on the single generated output. We report only the Top-10 results in the main table, while the Top-1, Top-3, and Top-5 metrics are provided in Appendix \ref{app: Supplementary}.

\noindent{\textbf{Empirical Results.}} 
Table \ref{tab:main_operation} summarizes the results. 
Overall, LLM-based baselines exhibit limited performance in the SOP setting. We observe no consistent improvement with increasing model size; instead, smaller models such as GPT-4o-mini achieve relatively better performance compared to larger LLMs. This suggests that extensive general knowledge about target materials may act as noisy information when generating structured synthesis operation sequences, in the absence of an appropriate conditioning or decision strategy, rather than providing a clear benefit.
SPENDE~\cite{na2023artificial}, which is specifically designed for thermoelectric materials, shows low performance on more general materials, indicating that it is not well-suited for SOP prediction in broad chemical spaces. 
The strongest baseline performance is obtained when retrieval-based methods \cite{he2023precursor} are employed, which more closely resemble the workflow of experimental chemists. Although this approach outperforms other models, it remains heavily dependent on the retrieved synthesis procedures of similar materials and therefore exhibits limited predictive generalization.
By contrast, the proposed method consistently and substantially outperforms all baselines across all splits and evaluation metrics. This improvement can be attributed to the material group decision chain, which enables the model to select synthesis strategies tailored to specific material classes, as well as to the use of explicit conditioning that allows precursor information to be effectively leveraged during synthesis operation prediction.

\subsection{Material Synthesis Planning (MSP) Task}
\label{ex_msp}
\begin{table*}[htbp] %
\centering
\small

\begingroup
\setlength{\aboverulesep}{0pt}
\setlength{\belowrulesep}{0pt}
\renewcommand{\arraystretch}{1.1}

\begin{minipage}[t]{0.58\textwidth}
    \vspace{0pt}
    \centering
    \captionof{table}{MSP Performance on two dataset splits.}
    \label{tab:main_msp}
    
    \resizebox{\linewidth}{!}{%
    \begin{tabular}{l cccc c cccc}
    \toprule
    \textbf{Method} & \multicolumn{4}{c}{\textbf{Split 1}} & & \multicolumn{4}{c}{\textbf{Split 2}} \\
    \cmidrule(lr){2-5} \cmidrule(lr){7-10}
    & \textbf{Top-1} & \textbf{Top-3} & \textbf{Top-5} & \textbf{Top-10} & & \textbf{Top-1} & \textbf{Top-3} & \textbf{Top-5} & \textbf{Top-10} \\
    \midrule
    
    \rowcolor{gray!12}
    \multicolumn{5}{l}{\textit{Large Language Models (API, few-shot prompting)}} & & \multicolumn{4}{l}{} \\
    \midrule
    GPT-4o-mini (20-shot) & 0.0 & 0.09 & 0.09 & 0.19 & & 0.19 & 0.28 & 0.28 & 0.47 \\
    GPT-4o-mini (40-shot) & 0.0 & 0.0 & 0.09 & 0.09 & & 0.27 & 0.27 & 0.37 & 0.64 \\
    GPT-4o (20-shot) & 0.27 & 0.55 & 0.64 & 0.73 & & 0.47 & 0.47 & 0.47 & 0.56 \\
    GPT-4o (40-shot) & 0.27 & 0.27 & 0.37 & 0.64 & & 0.37 & 0.84 & 0.93 & 1.12 \\
    GPT-5.1 (20-shot) & 0.46 & 0.64 & 0.73 & 1.01 & & 0.65 & 0.75 & 0.75 & 0.75 \\
    GPT-5.1 (40-shot) & 0.55 & 0.64 & 0.64 & 0.73 & & 0.37 & 0.37 & 0.47 & 0.47 \\
    Claude Haiku 4.5 (20-shot) & 0.56 & 0.65 & 0.65 & 0.74 & & 0.56 & 0.65 & 0.65 & 0.75 \\
    Claude Haiku 4.5 (40-shot) & 0.56 & 0.65 & 0.65 & 0.74 & & 0.37 & 0.84 & 0.93 & 1.12 \\
    GPT-5.1(\text{PP}) + GPT-4o-mini (\text{SOP}) & 0.91 & 1.65 & 2.01 & 3.75 & & 1.4 & 2.71 & 3.64 & 5.41 \\
    
    \midrule\midrule
    \textbf{\proposed~(Qwen-2.5-7B)} & \textbf{5.76} & \textbf{10.79} & \textbf{13.62} & \textbf{18.56} & & \underline{6.62} & \textbf{13.34} & \textbf{16.79} & \underline{22.48} \\
    \textbf{\proposed~(LLaMA-3.1-8B)} & \underline{4.75} & \underline{9.69} & \underline{13.35} & \underline{18.19} & & \textbf{6.72} & \underline{12.12} & \underline{16.60} & \textbf{23.23} \\
    \bottomrule
    \end{tabular}
    }
\end{minipage}
\hfill %
\begin{minipage}[t]{0.38\textwidth}
    \vspace{0pt}
    \centering
    \includegraphics[width=\linewidth]{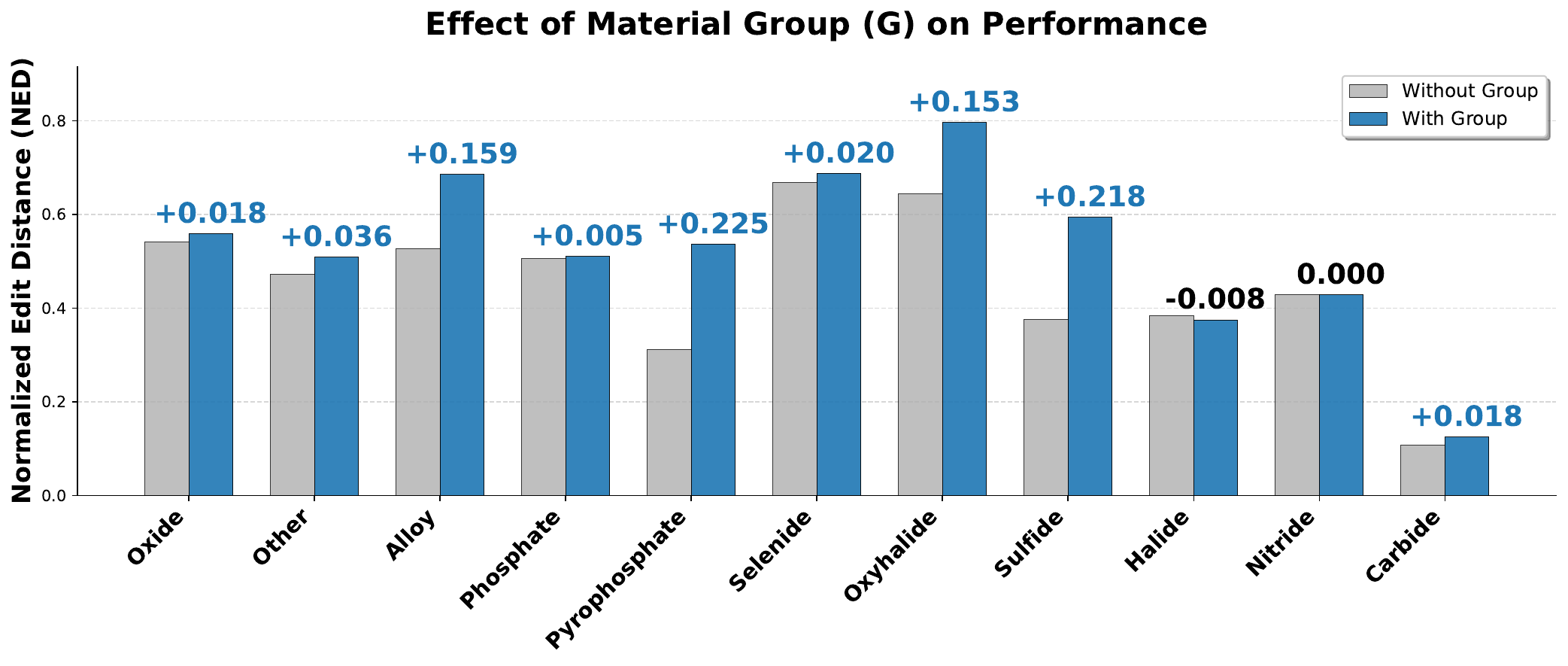}
    \captionof{figure}{Effect of the material group decision chain on performance (NED) across 11 material classes.}
    \label{fig:material_group_effect}
\end{minipage}

\endgroup
\end{table*}

\noindent{\textbf{Baseline Methods.}} Since no existing methods directly address the MSP task, we construct several baseline pipelines. For LLM-based baselines, we prompt a single LLM to perform the entire MSP task in a single process. Additionally, we combine the best-performing LLMs for each subtask into a hybrid pipeline for comparison. 
Note that \proposed~ employs LLMs that are independently fine-tuned for the PP and SOP task.

\noindent{\textbf{Evaluation Protocol.}} To evaluate MSP, both the predicted precursor set and synthesis operation sequence must be correct. For a rigorous assessment, we adopt the Top-$K$ Exact Match metric, which counts a prediction as correct only when both subtasks, PP and SOP, are simultaneously matched to the ground truth. For both our \proposed~ and the LLM-based hybrid pipeline, SOP is conditioned on the top-1 precursor prediction during evaluation.

\noindent{\textbf{Empirical Results for MSP.}} Table \ref{tab:main_msp} summarizes the results. All LLM-based baselines struggle on MSP. With the exception of the hybrid pipeline (GPT-5.1 + GPT-4o-mini; each performs the best on PP and SOP, respectively), none of the methods exceeds 2\% in terms of Top-10 accuracy. This indicates that predicting both precursors and synthesis operations jointly with a single model is highly challenging, and that a two-stage pipeline can provide some benefit. Nevertheless, even the hybrid pipeline achieves only 3.75\% and 5.41\% on the two splits, respectively. Despite this difficulty, the proposed method attains substantially higher performance, achieving Top-10 scores of 18.56\% on Split~1 and 23.23\% on Split~2. Since this strict criterion may penalize chemically valid alternative routes, we additionally provide a relaxed evaluation in Appendix~\ref{app:relaxed_msp}, a chemical validity evaluation in Appendix~\ref{app:chemical_validity}, an error propagation analysis of the MSP pipeline in Appendix~\ref{app:error_decomp}, extended comparisons with additional baselines in Appendix~\ref{app:extended_api} and per-class performance in Appendix~\ref{app:per_class}.

\subsection{Further Analyses}

\noindent{\textbf{Ablation Studies.}} To evaluate the contribution of each component in the MSP subtasks, we perform ablation studies in Tables \ref{tab:ablation} (a) and (b). 
For both PP and SOP, removing the material group decision chain leads to a noticeable performance degradation, highlighting the importance of structured decision-making for both tasks. For SOP, we further examine the effects of explicit conditioning and the precursor type information. We observe that explicit conditioning and the precursor types play a critical role in the observed performance gains, underscoring the importance of incorporating precursor constraints in synthesis operation prediction. 

\begingroup
\setlength{\intextsep}{0pt}
\begin{wraptable}{r}{0.50\textwidth}
\centering
\captionsetup{position=top,skip=7pt,belowskip=0pt}
\caption{Ablation studies on PP and SOP (Split~1). EC: explicit conditioning with precursor types.}
\label{tab:ablation}
\setlength{\tabcolsep}{3pt}
\renewcommand{\arraystretch}{1.0}
\resizebox{\linewidth}{!}{%
\begin{tabular}{l cc cccc}
\toprule
\multicolumn{7}{c}{\textbf{(a) Precursor Prediction (PP)}} \\
\midrule
\textbf{Model} & \multicolumn{2}{c}{\textbf{$G$}} & \textbf{Top-1} & \textbf{Top-3} & \textbf{Top-5} & \textbf{Top-10} \\
\midrule
(A) Naive Fine-tuning & \multicolumn{2}{c}{} & 70.59 & 84.90 & 90.62 & 94.39 \\
\midrule
(B) + Material Group \textbf{(\proposed)} & \multicolumn{2}{c}{\textbf{\checkmark}} & \textbf{70.75} & \textbf{85.56} & \textbf{91.68} & \textbf{94.97} \\
\bottomrule
\addlinespace[5pt]
\toprule
\multicolumn{7}{c}{\textbf{(b) Synthesis Operation Prediction (SOP)}} \\
\midrule
\textbf{Model} & \textbf{$G$} & \textbf{EC} & \textbf{Top-10} & \textbf{NED}$\uparrow$ & \textbf{LCS}$\uparrow$ & \textbf{F-1}$\uparrow$ \\
\midrule
(A) Naive Fine-tuning              &            &            & 11.24 & 0.5336 & 0.6311 & 0.6572 \\
(B) + Material Group               & \checkmark &            & 13.35 & 0.5623 & 0.6605 & 0.6864 \\
(C) + Explicit Conditioning        &            & \checkmark & 25.13 & 0.7168 & 0.8064 & 0.8342 \\
(D) $G$ + EC w/o precursor types   & \checkmark & \checkmark & 25.50 & 0.7188 & 0.8067 & 0.8353 \\
\midrule
\textbf{(E) \proposed~(Qwen-2.5-7B)} & \textbf{\checkmark} & \textbf{\checkmark} & \textbf{26.51} & \textbf{0.7207} & \textbf{0.8100} & \textbf{0.8359} \\
\bottomrule
\end{tabular}}
\end{wraptable}

Moreover, to examine the effectiveness of the material group decision chain, we analyze how it improves performance across the 11 material classes. As shown in Figure~\ref{fig:material_group_effect}, simply introducing the decision chain yields consistent performance gains in the majority of classes (9 out of 11). Notably, although oxides constitute approximately 85\% of the dataset, the decision chain provides chemically plausible guidance across different material categories, rather than biasing predictions toward the dominant class. The group classification performance is reported in Appendix~\ref{app:group_classification}.

\par
\endgroup

\noindent{\textbf{Integration with Downstream Condition Prediction Models.}} As \proposed~produces both precursors and synthesis operations, a publicly available condition prediction model~\citep{huo2022machine} that takes this information as input can be naturally combined with our framework to additionally predict temperature and duration for heating operations. On the target-disjoint split (Split~2), the integrated pipeline achieves a Mean Absolute Error (MAE) of $134.9$\,\textdegree C and $6.7$\,h for heating operations. 
Table~\ref{tab:case_studies} presents cases in which \proposed~successfully predicts the precursor set and synthesis operation sequence, and the specific heating conditions are generated via the integrated framework. Overall, the resulting plan provides an informative starting point for synthesis planning of target materials. Additional cases are provided in Appendix~\ref{app:additional_case_studies}.
\begin{table}[!htbp]
\centering
\captionsetup{position=top,skip=7pt,belowskip=0pt}
\caption{Synthesis plan generation on unseen materials. Cells show \textbf{predicted}\,/\,reported values from \proposed~integrated with the condition model~\citep{huo2022machine}. DOIs in Appendix.}
\label{tab:case_studies}
\scriptsize
\setlength{\tabcolsep}{2pt}
\renewcommand{\arraystretch}{0.7}
\definecolor{caseheader}{RGB}{230, 240, 250}
\resizebox{0.70\textwidth}{!}{%
\begin{tabular}{@{}clcc|clcc@{}}
\toprule
\textbf{\#} & \textbf{Op.} & \textbf{$T$ (P/R, \textdegree C)} & \textbf{$t$ (P/R, h)} & \textbf{\#} & \textbf{Op.} & \textbf{$T$ (P/R, \textdegree C)} & \textbf{$t$ (P/R, h)} \\
\midrule
\rowcolor{caseheader}\multicolumn{4}{@{}l|}{\textbf{Case 1:} Ba$_{0.774}$Ca$_{0.226}$Ti$_{0.9}$Sn$_{0.1}$O$_3$} & \multicolumn{4}{l}{\textbf{Case 2:} Nd$_{0.7}$Sr$_{0.3}$Mn$_{0.9}$Ga$_{0.1}$O$_3$} \\
\rowcolor{caseheader}\multicolumn{4}{@{}l|}{\textit{Pre.:} BaCO$_3$, CaCO$_3$, SnO$_2$, TiO$_2$} & \multicolumn{4}{l}{\textit{Pre.:} Ga$_2$O$_3$, MnO$_2$, Nd$_2$O$_3$, SrCO$_3$} \\
\rowcolor{caseheader}\multicolumn{4}{@{}l|}{\textit{Ops:} mix$\to$dry$\to$\textbf{heat}$\to$mix$\to$shape$\to$\textbf{sinter}} & \multicolumn{4}{l}{\textit{Ops:} mix$\to$\textbf{heat}$\to$mix$\to$\textbf{heat}$\to$shape$\to$\textbf{sinter}} \\
\cmidrule(lr){1-4}\cmidrule(lr){5-8}
0 & heat   & \textbf{1108}\,/\,1200 & \textbf{3.9}\,/\,4.0  & 0 & heat   & \textbf{1066}\,/\,1000 & \textbf{24.3}\,/\,24.0 \\
1 & sinter & \textbf{1347}\,/\,1400 & \textbf{3.4}\,/\,4.0  & 1 & heat   & \textbf{1173}\,/\,1350 & \textbf{24.7}\,/\,24.0 \\
  &        &                        &                       & 2 & sinter & \textbf{1398}\,/\,1350 & \textbf{22.4}\,/\,24.0 \\
\bottomrule
\end{tabular}}
\end{table}

\section{Related Works}
\label{sec:related}
\noindent\textbf{Material Synthesis Planning.}
MSP is a long-standing challenge in chemical sciences~\citep{mat_sci_auto_lab,msp_chl2}. Conventional approaches based on DFT~\citep{dft,dft_sdl} require atomic structures of products and precursors, which are typically unavailable in real experiments~\citep{msp_inorg,llm_msp2}. To bypass this requirement, recent machine learning methods for composition-based MSP have focused on either predicting precursor sets~\citep{he2023precursor,noh2024retrieval} or generating sequences of synthesis operations~\citep{na2023artificial}. However, no existing method addresses both subtasks within a single framework, leaving the MSP task unresolved.

\noindent\textbf{LLMs for Materials Science.}
LLMs have been adopted in materials science to leverage their prior knowledge of physical chemistry, supporting downstream predictive and generative models~\citep{llm_mat_sci1,llm_mat_sci2,llm_mat_prop_pred,llm_polymer_prop_pred}. Building on advances in reasoning capability, recent works have begun to address MSP. RHS2LHS and TGT2CEQ~\citep{llm_msp1} infer precursors and chemical reactions, but require ground-truth synthesis pathways as additional input and do not address SOP. LLM-QDMD~\citep{llm_msp2} designs experiments for quantum dot synthesis, but is restricted to refining existing plans within a single material domain. A general LLM-based framework that performs both precursor and operation prediction across diverse materials remains unexplored. A more detailed discussion of related work is provided in Appendix~\ref{app: Extended related works}.

\section{Conclusion}
\label{sec:conclusion}
This paper proposes \proposed, an LLM-based framework for MSP that decomposes the task into two subproblems: precursor prediction (PP) and synthesis operation prediction (SOP), each addressed through structured fine-tuning with a material group decision chain and explicit precursor conditioning. Our framework achieves 94.97\% Top-10 accuracy on PP, an F-1 score of 0.8359 on SOP, and consistently outperforms baselines on the MSP task. We further show that integrating the framework with a condition prediction model yields synthesis plans including temperature and duration, providing an informative starting point for synthesis planning of target materials. These results highlight the practical potential of \proposed~for autonomous MSP, a long-standing challenge in chemical sciences.

\section*{Acknowledgement}
This work was supported by the National Research Foundation of Korea (NRF) grant funded by the Korea government (MSIT) (RS-2026-25607361). This research was also supported by the Fundamental Research Program (PNKB300) of the Korea Institute of Materials Science, funded by the Ministry of Science and ICT. Additional support was provided by a National Research Foundation of Korea (NRF) grant funded by the Ministry of Science and ICT (RS-2022-NR068758).

\clearpage

\linespread{0.98}\selectfont
\bibliographystyle{neurips_2026}
\bibliography{neurips_2026}

\clearpage

\appendix
\begin{center}
    \huge{\emph{Supplementary Material}}
\end{center}
\begin{center}
    \large{\emph{- MSP-LLM: A Structured Large Language Model Framework for Inorganic Material Synthesis Planning -}}
\end{center}
\vspace*{10mm}

\DoToC

\clearpage

\section{Extended Related Works}
\label{app: Extended related works}
\noindent\textbf{Material Synthesis Planning. }
MSP is a long-standing challenge in chemical sciences due to the inherent complexity and uncertainty associated with chemical reactions \cite{mat_sci_auto_lab,msp_chl2}. In materials science, calculation- and simulation-based approaches on physical chemistry, often grounded in density functional theory (DFT) \cite{dft}, have been used to screen precursor candidates and to analyze thermodynamically stable synthesis pathways toward target product materials \cite{dft_sdl}. However, these conventional methods have faced a fundamental limitation in real-world applications: the atomic structures of product and candidate precursor materials, which are required for reliable first-principles calculations, are typically unavailable in most chemical experiments \cite{msp_inorg,llm_msp2}. 
To overcome this challenge, prior machine learning methods for composition-based MSP have focused on either predicting precursor sets \cite{he2023precursor,noh2024retrieval} or generating sequences of synthesis operations \cite{na2023artificial}. 
However, MSP remains unresolved because there is currently no computational method capable of solving the MSP task, which requires predicting both plausible precursors and corresponding synthesis pathways.

\noindent\textbf{LLM for Materials Science. }
LLMs have been increasingly adopted in materials science to exploit their broad prior knowledge of physical chemistry, thereby supporting the construction of downstream predictive and generative models \cite{llm_mat_sci1,llm_mat_sci2}. In particular, LLMs have notably accelerated materials discovery by facilitating efficient high-throughput screening based on surrogate models that predict the physical and chemical properties of inorganic materials \cite{llm_mat_prop_pred} and polymers \cite{llm_polymer_prop_pred}. With the advances in the reasoning capabilities of LLMs, they have recently emerged as a method for addressing the MSP task, which constitutes a next remaining challenge for materials discovery. In this direction, RHS2LHS and TGT2CEQ were proposed to infer precursor materials and chemical reactions for designing an initial stage of the material synthesis plan \cite{llm_msp1}. However, both RHS2LHS and TGT2CEQ require additional information about ground-truth synthesis pathways when predicting precursors and do not address the SOP tasks. LLM-QDMD is an LLM-driven MSP method for designing experiment plans for the synthesis of quantum dot materials \cite{llm_msp2}. Although LLM-QDMD achieved substantial success within the domain of quantum dot materials, its applicability to real-world applications is inherently limited because it focuses on refining existing synthesis plans for quantum dot materials rather than generating new synthesis plans across general materials spaces.

\section{Implementation Details}
\label{app: Implementation}
Implementation details of \proposed~ are provided in this section. 

\subsection{Training Details}
All models are fine-tuned using Python 3.10.18, PyTorch 2.3.1, Accelerate 1.10.1, and Hugging Face Hub 0.34.4 on 2 NVIDIA RTX A6000 GPUs. To accommodate resource-limited settings, we employ QLoRA for parameter-efficient fine-tuning. The training hyperparameter configurations for each task and LLM are reported in Table \ref{tab:hyperparams}

\begin{table}[hbtp]
\centering
\caption{Hyperparameter configuration for PP and SOP across different LLM backbones and dataset splits.}
\label{tab:hyperparams}
\resizebox{\columnwidth}{!}{%
\begin{tabular}{p{3.5cm} l c c c c c c c} %
\hline
\textbf{LLM} & \textbf{Task} 
& \multicolumn{3}{c}{\textbf{Split 1}} 
& %
& \multicolumn{3}{c}{\textbf{Split 2}} \\
\cline{3-5} \cline{7-9} %
 &  & \textbf{LoRA Rank ($r$)} & \textbf{LoRA Alpha} & \textbf{lr / bs} 
 & %
 & \textbf{LoRA Rank ($r$)} & \textbf{LoRA Alpha} & \textbf{lr / bs} \\
\hline
LLaMA-3.1-8B\\ \cite{grattafiori2024llama} 
& PP  
& 16 & 32 & 1e-4 / 4 
& %
& 16 & 32 & 1e-4  / 2 \\

LLaMA-3.1-8B\\ \cite{grattafiori2024llama} & SOP 
& 16 & 32 & 1e-4  / 2 
& %
& 16 & 32 & 1e-4 / 2 \\

Qwen-2.5-7B\\ \cite{qwen2025qwen25technicalreport} & PP  
& 16 & 32 & 1e-4 / 4 
& %
& 16 & 32 & 1e-4  / 4 \\

Qwen-2.5-7B\\ \cite{qwen2025qwen25technicalreport} & SOP 
& 16 & 32 & 1e-4  / 2 
& %
& 16 & 32 & 1e-4  / 4 \\
\hline
\end{tabular}%
}
\end{table}

\section{Dataset}
\label{app: Dataset}
\subsection{Preprocessing}
\label{app:preprocessing}
We preprocess the real-world experimental inorganic synthesis dataset \cite{kononova2019text}.

\noindent{\textbf{Precursor Prediction (PP).}} 
We filtered out samples that do not contain synthesis operation–related information and removed data associated with compositional systems. We further excluded cases in which the corresponding precursor set contained only a single precursor. In addition, we applied a frequency-based filtering over the full dataset, retaining only precursors that appear at least five times. As a result, 288 unique precursors remained in the final dataset.

\noindent{\textbf{Synthesis Operation Prediction (SOP). }} In addition to the five basic synthesis operations provided in the original dataset—heating, mixing, shaping, quenching, and drying—we refine the heating category to better reflect real-world experimental practice. Specifically, we introduce two subcategories, sintering and annealing, and classify the remaining cases as generic heating. To distinguish these categories, we leverage synthesis-related subkeywords in the dataset: samples containing subkeywords related to “sinter” are labeled as sintering, while those containing subkeywords related to “anneal” are labeled as annealing.

To capture the synthesis intent and strategy of experimentalists, as well as coarse-grained material information (e.g., thin films), we use the titles of the source literature as contextual cues (synthesis context). We extract the following categories from the titles using GPT-4o:
'host material', 'dopant or substitution', 'material class', 'functional property', 'composition control', and 'processing or stimulus'. When relevant information is not present, the corresponding field is set to None. Since 'processing or stimulus' contains information directly related to synthesis operations, we exclude this field from the model inputs to ensure a precise evaluation. Instead, only the extracted information from 'host material', 'dopant or substitution', 'material class', 'functional property', and 'composition control', is used by all baselines as well as \proposed.

\subsection{Dataset Split}

Fundamentally, a single material can be synthesized using multiple precursors and multiple operations. In other words, a single chemical formula can correspond to multiple synthesis routes, which implies a one-to-many relationship. Through data preprocessing, we removed duplicates and kept only unique combinations of chemical formula–precursor–operation.

In Split 1, the dataset was randomly divided into train/validation/test sets with an 8:1:1 ratio, without restricting chemical formulas. As a result, the same chemical formula may appear in the train, validation, and test sets. However, because the precursors and operations that must be predicted in the PP and SOP tasks differ from those seen during training, the model still needs to infer different precursors and operations even for a chemical formula it has already encountered.

In Split 2, the dataset was also divided with an 8:1:1 ratio, but chemical formulas were enforced to be non-overlapping across the train, validation, and test sets. This setup allows evaluation of the model’s ability to perform PP and SOP on chemical formulas that were completely unseen during training, thereby measuring its generalization capability with respect to chemical formulas.

\section{Baselines}
\label{app: Baselines}
\noindent{\textbf{LLM Baselines.}}
We compare LLM baselines including GPT-4o-mini\footnote{\url{https://openai.com/index/gpt-4o-mini-advancing-cost-efficient-intelligence/}}
, GPT-4o\footnote{\url{https://openai.com/index/hello-gpt-4o/}}
, GPT-5.1\footnote{\url{https://openai.com/index/gpt-5-1/}}
, and Claude Haiku 4.5\footnote{\url{https://www.anthropic.com/claude/haiku}}
. All models are accessed via their respective APIs. The GPT models are configured with a temperature of 0.8, while Claude Haiku is set to 0.3.

\noindent{\textbf{Specialized Models for Precursor Prediction (PP).}} We compared our approach with a retriever-based method \citet{he2023precursor} that learns the relationship between target materials and precursors through masked precursor completion, as well as a more recent work, Retrieval Retro\citep{noh2024retrieval}, which employs an advanced attention mechanism and incorporates thermodynamic relationships between targets and precursors to identify more feasible precursor candidates. For both methods, we used their official implementations available on GitHub. In addition, when solving PP with the LLM baselines, we directly used the prompt from \cite{prein2025language}.

\noindent{\textbf{Specialized Models for Precursor Prediction (SOP).}} Using the retriever from \cite{he2023precursor} that was employed for PP, we leveraged the synthesis routes of materials similar to the target material as the predicted outputs, serving as a baseline that reflects the practices of real synthesis experts. In addition, we adopted SPENDE\citep{na2023artificial} which is specifically designed for thermoelectric material synthesis prediction, using the official implementation provided on GitHub. The prompt used by the LLM baselines for the SOP task is provided in the Appendix \ref{app: prompt}.

\section{Additional Experiments}
\label{app: Additional Experiments}
\subsection{Synthesis Context Usage}
\label{app:synthesis_context}
We examine the contribution of synthesis context on Split~2 by comparing three settings: context omitted during both training and testing, omitted only at test time, and included during both. As described in Appendix~\ref{app: Dataset}, context comprises five attributes extracted from paper titles; the operation-related ``processing or stimulus'' field is excluded. PP uses the target composition, while SOP additionally receives precursors and, when available, context. Standalone SOP uses ground-truth precursors; MSP uses the Top-1 PP prediction. Thus, context-free MSP requires only the target composition as external input.

\begin{table}[htbp]
\centering
\captionsetup{position=top,skip=7pt}
\caption{Effect of synthesis context on Split~2. Yes/No indicates whether context is provided during training or testing. Top-$K$ exact-match scores are percentages.}
\label{tab:synthesis_context_usage}
\small
\setlength{\tabcolsep}{5pt}
\renewcommand{\arraystretch}{1.12}
\begin{tabular}{cc cccc cccc}
\toprule
\multicolumn{2}{c}{\textbf{Context}} & \multicolumn{4}{c}{\textbf{Qwen-2.5-7B}} & \multicolumn{4}{c}{\textbf{LLaMA-3.1-8B}} \\
\cmidrule(lr){1-2}\cmidrule(lr){3-6}\cmidrule(lr){7-10}
Train & Test & \multicolumn{8}{c}{\textbf{(a) Synthesis Operation Prediction (SOP)}} \\
& & Top-10 & NED$\uparrow$ & LCS$\uparrow$ & F-1$\uparrow$ & Top-10 & NED$\uparrow$ & LCS$\uparrow$ & F-1$\uparrow$ \\
\midrule
No & No & 29.94 & .732 & .816 & .843 & 29.38 & .728 & .816 & .844 \\
Yes & No & 29.66 & .730 & .815 & .839 & 28.92 & .726 & .812 & .836 \\
Yes & Yes & 32.18 & .742 & .824 & .849 & 31.44 & .742 & .824 & .850 \\
\midrule
& & \multicolumn{8}{c}{\textbf{(b) Material Synthesis Planning (MSP)}} \\
& & Top-1 & Top-3 & Top-5 & Top-10 & Top-1 & Top-3 & Top-5 & Top-10 \\
\midrule
No & No & 6.81 & 12.31 & 15.11 & 20.71 & 5.97 & 11.29 & 14.65 & 20.06 \\
Yes & No & 6.53 & 11.29 & 14.46 & 20.06 & 5.13 & 11.19 & 15.39 & 20.15 \\
Yes & Yes & 6.62 & 13.34 & 16.79 & 22.48 & 6.72 & 12.12 & 16.60 & 23.23 \\
\bottomrule
\end{tabular}
\end{table}

Table~\ref{tab:synthesis_context_usage} shows that removing context from both training and testing reduces SOP Top-10 by 2.24 and 2.06 percentage points for Qwen and LLaMA, respectively. MSP Top-10 remains 20.71\% and 20.06\%, compared with 22.48\% and 23.23\% with context. Context therefore provides useful information, but is not required to generate synthesis plans. This limited sensitivity is consistent with the nature of the context: the extracted fields are already sparse, with 23.3\%, 22.1\%, and 22.8\% missing in the training, validation, and test sets (up to 38.2\% for ``Dopant or substitution''), and they are only indirectly related to synthesis operations.

\subsection{Error Propagation Analysis in the \proposed~ Pipeline }
\label{app:error_decomp}
\begin{table}[!htbp]
\centering
\caption{Error propagation analysis on Split~2 (LLaMA-3.1-8B). (\textit{Left}) Comparison between \proposed~and an oracle SOP that always uses ground-truth precursors as input. (\textit{Right}) Decomposition of MSP errors into four cases based on PP and SOP correctness.}
\label{tab:error_propagation}

\begin{minipage}[t]{0.42\textwidth}
\vspace{0pt}
\centering
\subcaption{Comparison with Oracle SOP.}
\label{tab:oracle_comparison}
\footnotesize                            %
\setlength{\tabcolsep}{4pt}              %
\begingroup
\setlength{\aboverulesep}{0pt}
\setlength{\belowrulesep}{0pt}
\renewcommand{\arraystretch}{1.2}
\begin{tabular}{l cccc}
\toprule
 & \textbf{Top-1} & \textbf{Top-3} & \textbf{Top-5} & \textbf{Top-10} \\
\midrule
\proposed    & 6.72  & 12.12 & 16.60 & 23.23 \\
Oracle SOP & 9.20  & 16.62 & 22.73 & 31.80 \\
\midrule
Gap        & 2.48\,pp & 4.50\,pp & 6.13\,pp & 8.57\,pp \\
\bottomrule
\end{tabular}
\endgroup
\end{minipage}%
\hfill
\begin{minipage}[t]{0.55\textwidth}
\vspace{0pt}
\centering
\subcaption{Error decomposition (Top-10).}
\label{tab:error_decomposition}
\footnotesize                            %
\setlength{\tabcolsep}{4pt}
\begingroup
\setlength{\aboverulesep}{0pt}
\setlength{\belowrulesep}{0pt}
\renewcommand{\arraystretch}{1.2}
\begin{tabular}{l cc}
\toprule
\textbf{Case} & \textbf{Count} & \textbf{Ratio} \\
\midrule
A: PP\cmark~SOP\cmark~(MSP success)    & 249 & 23.2\% \\
B: PP\cmark~SOP\xmark~(SOP bottleneck) & 534 & 49.8\% \\
C: PP\xmark~SOP\cmark~(PP bottleneck)  & 89  & 8.3\%  \\
D: PP\xmark~SOP\xmark~(both failed)    & 200 & 18.7\% \\
\bottomrule
\end{tabular}
\endgroup
\end{minipage}
\end{table}

To understand where errors arise in the \proposed~pipeline, we conduct two complementary analyses. First, Table~\ref{tab:error_propagation} compares \proposed~with an oracle SOP that always uses ground-truth precursors as input; the resulting Top-10 gap of $8.57$\,pp quantifies the contribution of PP errors to overall MSP failure. Second, Table~\ref{tab:error_decomposition} decomposes MSP failures into four cases based on PP and SOP correctness. Case~B (PP correct but SOP fails) accounts for the largest share at $49.8\%$, indicating that SOP is the primary bottleneck of the joint pipeline. Together, these results suggest that improving SOP offers the largest opportunity for advancing MSP, while reducing PP errors provides a complementary $8.57$\,pp room for further gain.

\subsection{Further Baseline Comparisons}
\label{app:extended_api}
\begin{table*}[h]
\centering
\caption{PP, SOP, and MSP performance on Split~2.}
\label{tab:main_combined_split2}
\small
\begingroup
\setlength{\aboverulesep}{0pt}
\setlength{\belowrulesep}{0pt}
\renewcommand{\arraystretch}{1.3}
\resizebox{\textwidth}{!}{%
\begin{tabular}{l cccc c cccc c cccc}
\toprule
& \multicolumn{4}{c}{\textbf{PP}} & & \multicolumn{4}{c}{\textbf{SOP}} & & \multicolumn{4}{c}{\textbf{MSP}} \\
\cmidrule(lr){2-5} \cmidrule(lr){7-10} \cmidrule(lr){12-15}
\textbf{Method}
& \textbf{Top-1} & \textbf{Top-3} & \textbf{Top-5} & \textbf{Top-10}
& & \textbf{Top-10} & \textbf{NED}$\uparrow$ & \textbf{LCS}$\uparrow$ & \textbf{F-1}$\uparrow$
& & \textbf{Top-1} & \textbf{Top-3} & \textbf{Top-5} & \textbf{Top-10} \\
\midrule

\rowcolor{gray!12}
\multicolumn{15}{l}{\textit{Large Language Models (API, few-shot prompting)}} \\
\midrule
GPT-4o-mini (20-shot)
& 54.85 & 63.40 & 67.91 & 70.34
& & 7.09  & 0.5847 & 0.7089 & 0.7407
& & 0.19  & 0.28   & 0.28   & 0.47   \\
GPT-4o-mini (40-shot)
& 55.97 & 64.27 & 67.82 & 71.36
& & 7.93  & 0.5934 & 0.7141 & 0.7442
& & 0.27  & 0.27   & 0.37   & 0.64   \\
GPT-4o-mini (60-shot)
& 55.69 & 65.02 & 68.75 & 72.20
& & 10.54 & 0.6099 & 0.7310 & 0.7616
& & 1.40  & 1.96   & 2.05   & 2.24   \\
GPT-4o-mini (80-shot)
& 59.51 & 67.44 & 71.92 & 75.37
& & 9.24  & 0.6048 & 0.7238 & 0.7551
& & 0.93  & 1.21   & 1.31   & 1.40   \\
GPT-4o-mini (100-shot)
& 60.54 & 68.28 & 71.64 & 74.35
& & 9.70  & 0.6082 & 0.7274 & 0.7572
& & 0.93  & 1.59   & 1.59   & 1.68   \\
GPT-4o (20-shot)
& 60.17 & 69.31 & 73.04 & 76.31
& & 3.82  & 0.5670 & 0.6865 & 0.7166
& & 0.47  & 0.47   & 0.47   & 0.56   \\
GPT-4o (40-shot)
& 61.57 & 71.27 & 73.97 & 77.05
& & 4.94  & 0.5749 & 0.6924 & 0.7245
& & 0.37  & 0.84   & 0.93   & 1.12   \\
GPT-4o (60-shot)
& 62.78 & 72.20 & 75.09 & 77.05
& & 6.53  & 0.5888 & 0.7093 & 0.7398
& & 0.75  & 0.93   & 0.93   & 1.03   \\
GPT-4o (80-shot)
& 63.15 & 72.67 & 76.12 & 78.82
& & 6.81  & 0.5883 & 0.7098 & 0.7392
& & 0.47  & 0.93   & 1.03   & 1.12   \\
GPT-4o (100-shot)
& 63.72 & 71.46 & 75.28 & 78.36
& & 8.02  & 0.5926 & 0.7139 & 0.7430
& & 0.56  & 0.84   & 1.03   & 1.31   \\
\midrule

\rowcolor{gray!12}
\multicolumn{15}{l}{\textit{Large Language Models (API, fine-tuned)}} \\
\midrule
GPT-4o-mini (fine-tuning)
& 69.78 & 79.48 & 81.90 & 86.10
& & 15.11 & 0.6691 & 0.7727 & 0.8048
& & 1.77  & 4.20   & 6.72   & 12.03  \\

\midrule\midrule
\textbf{\proposed~(Qwen-2.5-7B)}
& \underline{71.92} & \underline{88.62} & \underline{92.72} & \underline{95.53}
& & \textbf{32.18} & \underline{0.7420} & \textbf{0.8240} & \underline{0.8493}
& & \underline{6.62} & \textbf{13.34} & \textbf{16.79} & \underline{22.48} \\
\textbf{\proposed~(LLaMA-3.1-8B)}
& \textbf{72.85} & \textbf{88.99} & \textbf{93.56} & \textbf{96.36}
& & \underline{31.44} & \textbf{0.7424} & \underline{0.8235} & \textbf{0.8498}
& & \textbf{6.72} & \underline{12.12} & \underline{16.60} & \textbf{23.23} \\
\bottomrule
\end{tabular}
}%
\endgroup
\vspace{-2ex}
\end{table*}

We further conduct additional experiments with GPT-4o-mini and GPT-4o on PP, SOP, and MSP tasks. We extend the few-shot prompting setup to 60, 80, and 100 shots, and additionally fine-tune GPT-4o-mini on the same training data using naive fine-tuning (without \proposed's task-specific strategies).

For few-shot prompting, performance slightly improves as the number of shots increases but quickly saturates and remains substantially below \proposed, while also introducing practical challenges such as expanded context windows and increased computational cost. For fine-tuning, while GPT-4o-mini shows substantial improvement over its few-shot counterpart, our framework still achieves consistently higher performance across all tasks and metrics, indicating that the advantage of \proposed~persists even against a fine-tuned proprietary LLM.

We compare three additional MSP baselines on Split~2: naive two-stage fine-tuning and end-to-end fine-tuning using the same LLaMA-3.1-8B backbone, and nearest-neighbor retrieval. All settings use synthesis context. The retrieval baseline identifies similar training instances using target composition and synthesis context with MatSciBERT~\citep{gupta2022matscibert}, and returns their precursor sets and operation sequences. These comparisons examine whether the gains arise merely from fine-tuning on the same data or retrieving similar synthesis routes.

\begin{table}[htbp]
\centering
\captionsetup{position=top,skip=7pt}
\caption{Further MSP baseline comparisons on Split~2 (exact match, \%). Fine-tuned models use LLaMA-3.1-8B; all settings use synthesis context.}
\label{tab:further_baselines}
\small
\setlength{\tabcolsep}{10pt}
\renewcommand{\arraystretch}{1.12}
\begin{tabular}{lcccc}
\toprule
\textbf{Method} & Top-1 & Top-3 & Top-5 & Top-10 \\
\midrule
Naive two-stage fine-tuning & 4.1 & 8.3 & 9.9 & 12.7 \\
End-to-end fine-tuning & 4.8 & 10.8 & 13.0 & 16.1 \\
Nearest-neighbor & 5.8 & 9.8 & 10.3 & 10.8 \\
\proposed & 6.72 & 12.12 & 16.60 & 23.23 \\
\bottomrule
\end{tabular}
\end{table}

Table~\ref{tab:further_baselines} shows that \proposed~outperforms these baselines at every cutoff, reaching 23.23\% Top-10 accuracy versus 12.7\%, 16.1\%, and 10.8\%, respectively. End-to-end training improves over naive two-stage fine-tuning, while retrieval gains little as the candidate budget increases. Since naive two-stage fine-tuning shares the backbone, training data, and two-stage decomposition with \proposed, the remaining gap reflects the contribution of the proposed structured fine-tuning strategies rather than fine-tuning on the same data alone.

\subsection{Chemical Validity Evaluation}
\label{app:chemical_validity}
We complement exact-match evaluation with two chemical-validity checks on predicted precursor sets. \emph{Element coverage} checks whether the precursors collectively contain the elements required by the target, while \emph{mass balance} checks whether the predicted precursor set permits a mass-balanced reaction. Table~\ref{tab:chemical_validity} reports both criteria at Top-1, Top-3, Top-5, and Top-10 on Split~2.

\begin{table}[htbp]
\centering
\captionsetup{position=top,skip=7pt}
\caption{Chemical validity evaluation of PP on Split~2. Values are pass rates (\%) for element coverage and mass balance.}
\label{tab:chemical_validity}
\small
\setlength{\tabcolsep}{5pt}
\renewcommand{\arraystretch}{1.12}
\begin{tabular}{lcccccccc}
\toprule
& \multicolumn{4}{c}{\textbf{Element coverage}} & \multicolumn{4}{c}{\textbf{Mass balance}} \\
\cmidrule(lr){2-5}\cmidrule(lr){6-9}
\textbf{Model} & @1 & @3 & @5 & @10 & @1 & @3 & @5 & @10 \\
\midrule
\proposed~(Qwen-2.5-7B) & 99.6 & 99.9 & 100.0 & 100.0 & 99.4 & 99.8 & 99.9 & 100.0 \\
\proposed~(LLaMA-3.1-8B) & 99.6 & 100.0 & 100.0 & 100.0 & 99.3 & 99.8 & 99.8 & 100.0 \\
\bottomrule
\end{tabular}
\end{table}

Both models achieve pass rates above 99\% at Top-1 and reach 100\% at Top-10. These results complement exact-match evaluation by assessing basic chemical consistency of the predicted precursors. Such checks are necessary constraints, however, and do not by themselves establish experimental feasibility or validate the predicted operation sequence.

\subsection{Relaxed MSP Performance of \proposed}
\label{app:relaxed_msp}
\begin{table}[!htbp]
\centering
\caption{Relaxed MSP performance of \proposed~(LLaMA-3.1-8B) on Split~2 with relaxed PP and SOP criteria. 
Percentages in parentheses denote the fraction of test samples whose predicted precursors pass the corresponding PP threshold.
PP Set-F1 measures precursor-level partial credit; PP Elem-F1 measures element-level coverage.}
\label{tab:relaxed_msp_pp}
\small
\begingroup
\setlength{\aboverulesep}{0pt}
\setlength{\belowrulesep}{0pt}
\renewcommand{\arraystretch}{1.15}
\setlength{\tabcolsep}{4pt}
\begin{tabular}{ll cccc}
\toprule
\textbf{PP Criterion} & \textbf{SOP Condition} &
\textbf{Top-1} & \textbf{Top-3} & \textbf{Top-5} & \textbf{Top-10} \\
\midrule
\rowcolor{gray!12}
\multicolumn{6}{l}{\textit{PP Set-F1 Criterion}} \\
\midrule
$\geq$0.8 ~(75.0\%) & SOP Exact           &  6.90 & 12.50 & 17.07 & 23.97 \\
$\geq$0.8            & SOP NED $\geq$0.8  & 12.31 & 20.24 & 25.00 & 32.74 \\
$\geq$0.8            & SOP LCS $\geq$0.8  & 17.82 & 30.88 & 38.06 & 49.35 \\
$\geq$0.8            & SOP F-1 $\geq$0.8  & \textbf{21.36} & \textbf{36.01} & \textbf{43.47} & \textbf{55.60} \\
\midrule
\rowcolor{gray!12}
\multicolumn{6}{l}{\textit{PP Elem-F1 Criterion}} \\
\midrule
$\geq$0.9 ~(91.2\%) & SOP Exact          &  7.56 & 15.58 & 21.27 & 29.38 \\
$\geq$0.9            & SOP NED $\geq$0.8  & 14.55 & 25.19 & 31.25 & 40.58 \\
$\geq$0.9            & SOP LCS $\geq$0.8  & 21.46 & 37.87 & 47.29 & 60.63 \\
$\geq$0.9            & SOP F-1 $\geq$0.8  & 25.65 & 43.75 & 53.36 & 67.91 \\
\cmidrule(lr){1-6}
$\geq$0.8 ~(99.4\%) & SOP Exact           &  8.12 & 16.51 & 22.57 & 31.44 \\
$\geq$0.8            & SOP NED $\geq$0.8  & 15.39 & 26.49 & 33.21 & 43.75 \\
$\geq$0.8            & SOP LCS $\geq$0.8  & 22.67 & 40.21 & 50.93 & 65.76 \\
$\geq$0.8            & SOP F-1 $\geq$0.8  & \textbf{27.05} & \textbf{46.36} & \textbf{57.37} & \textbf{73.13} \\
\bottomrule
\end{tabular}
\endgroup
\end{table}
The Top-$K$ exact-match metric used for MSP requires both the precursor set and the synthesis operation sequence to match the ground truth exactly, which can penalize chemically valid alternative routes. To allow partial credit for predictions that are chemically reasonable but not exact, we introduce two relaxed PP criteria and three relaxed SOP criteria. For PP, we consider (i) \textit{PP Set-F1}, which grants credit when the predicted precursor set has high overlap with the ground truth (F1 $\geq 0.8$), and (ii) \textit{PP Elem-F1}, which verifies whether the predicted precursors collectively supply all elements required by the target. For SOP, we apply soft thresholds: given an average ground-truth sequence length of $4.0$ steps, NED$\geq 0.8$ corresponds to at most one step difference, LCS$\geq 0.8$ to at least $3/4$ operations in correct order, and F1$\geq 0.8$ to at least $80\%$ operation overlap. These relaxed evaluations provide a chemically informed assessment of partial credit, highlighting \proposed's practical utility in providing synthesis guidance.

\clearpage
\suppressfloats[t]
\subsection{Per-class Performance Across PP, SOP, and MSP.}
\label{app:per_class}
\begin{table}[htbp!]
\centering
\caption{Per-class PP, SOP, and MSP performance (Split 1, LLaMA-3.1-8B).}
\label{tab:per_class_1}
\small
\begingroup
\setlength{\aboverulesep}{0pt}
\setlength{\belowrulesep}{0pt}
\renewcommand{\arraystretch}{1.3}
\resizebox{\columnwidth}{!}{%
\begin{tabular}{l r r r r r r r}
\toprule
\textbf{Class} & \textbf{n} & \textbf{PP Top-10} & \textbf{SOP Top-10} & \makecell{\textbf{MSP Top-10}\\(strict)} & \makecell{\textbf{MSP Top-10}\\(NED$\geq$0.8)} & \makecell{\textbf{MSP Top-10}\\(LCS$\geq$0.8)} & \makecell{\textbf{MSP Top-10}\\(F1$\geq$0.8)} \\
\midrule
oxide          & 938 & 95.3  & 25.59 & 18.0 & 29.00 & 45.52 & 51.28 \\
alloy          & 39  & 100.0 & 38.46 & 35.9 & 43.59 & 53.85 & 58.97 \\
phosphate      & 28  & 100.0 & 21.43 & 3.6  & 7.14  & 32.14 & 35.71 \\
pyrophosphate  & 14  & 92.9  & 28.57 & 14.3 & 14.29 & 21.43 & 21.43 \\
oxyhalide      & 9   & 66.7  & 44.44 & 11.1 & 11.11 & 22.22 & 22.22 \\
selenide       & 10  & 100.0 & 10.00 & 10.0 & 10.00 & 80.00 & 90.00 \\
sulfide        & 8   & 87.5  & 50.00 & 12.5 & 12.50 & 12.50 & 12.50 \\
not\_classified & 42  & 88.1  & 38.10 & 23.8 & 23.81 & 30.95 & 47.62 \\
carbide        & 1   & 100.0 & 0.00  & 0.0  & 0.00  & 100.0 & 100.0 \\
nitride        & 1   & 0.0   & 0.00  & 0.0  & 0.00  & 0.00  & 0.00 \\
halide         & 4   & 100.0 & 0.00  & 0.0  & 25.00 & 25.00 & 25.00 \\
\bottomrule
\end{tabular}
}%
\endgroup
\end{table}

\begin{table}[htbp!]
\centering
\caption{Per-class PP, SOP and MSP performance (Split 2, LLaMA-3.1-8B).}
\label{tab:per_class_2}
\small
\begingroup
\setlength{\aboverulesep}{0pt}
\setlength{\belowrulesep}{0pt}
\renewcommand{\arraystretch}{1.3}
\resizebox{\columnwidth}{!}{%
\begin{tabular}{l r r r r r r r}
\toprule
\textbf{Class} & \textbf{n} & \textbf{PP Top-10} & \textbf{SOP Top-10} & \makecell{\textbf{MSP Top-10}\\(strict)} & \makecell{\textbf{MSP Top-10}\\(NED$\geq$0.8)} & \makecell{\textbf{MSP Top-10}\\(LCS$\geq$0.8)} & \makecell{\textbf{MSP Top-10}\\(F1$\geq$0.8)} \\
\midrule
oxide          & 907 & 96.9  & 30.10 & 21.9 & 31.75 & 48.18 & 54.47 \\
alloy          & 61  & 100.0 & 45.90 & 42.6 & 42.62 & 52.46 & 57.38 \\
phosphate      & 28  & 85.7  & 21.43 & 3.6  & 7.14  & 21.43 & 28.57 \\
pyrophosphate  & 9   & 88.9  & 55.56 & 22.2 & 22.22 & 44.44 & 44.44 \\
oxyhalide      & 11  & 81.8  & 18.18 & 27.3 & 36.36 & 45.45 & 45.45 \\
selenide       & 7   & 100.0 & 42.86 & 42.9 & 42.86 & 57.14 & 71.43 \\
sulfide        & 12  & 100.0 & 50.00 & 50.0 & 50.0  & 58.33 & 75.0 \\
not\_classified & 32  & 87.5  & 37.50 & 21.9 & 21.88 & 40.62 & 46.88 \\
carbide        & 4   & 100.0 & 50.00 & 50.0 & 50.0  & 75.0  & 75.0 \\
nitride        & 1   & 100.0 & 0.00  & 0.0  & 0.0   & 100.0 & 100.0 \\
halide & 0 &-- & -- & -- & -- & -- & -- \\
\bottomrule
\end{tabular}
}%
\endgroup
\end{table}
Tables~\ref{tab:per_class_1} and~\ref{tab:per_class_2} report per-class performance of \proposed. Despite oxide dominance ($\sim$85\%), several minority classes achieve higher performance than oxide across PP, SOP, and MSP, indicating that the framework is not biased toward the majority class and generalizes effectively to diverse material groups.

\subsection{Material Group Classification Performance}
\label{app:group_classification}
\begin{table}[!htbp]
\centering
\caption{Task-level accuracy and F-1 score on two dataset splits (LLaMA-3.1-8B).}
\label{tab:task_accuracy}
\small
\begingroup
\setlength{\aboverulesep}{0pt}
\setlength{\belowrulesep}{0pt}
\renewcommand{\arraystretch}{1.2}
\begin{tabular}{l cc c cc}
\toprule
\textbf{Task} &
\multicolumn{2}{c}{\textbf{Split 1}} &
& \multicolumn{2}{c}{\textbf{Split 2}} \\
\cmidrule(lr){2-3} \cmidrule(lr){5-6}
& \textbf{Accuracy} & \textbf{F-1} &
& \textbf{Accuracy} & \textbf{F-1} \\
\midrule
PP  & 96.0 & 0.83 & & 96.9 & 0.80 \\
SOP & 95.4 & 0.72 & & 96.4 & 0.78 \\
\bottomrule
\end{tabular}
\endgroup
\end{table}

We report the material group classification performance of the decision chain in Table~\ref{tab:task_accuracy}. The classification accuracy of around 96\% and F-1 scores ranging from 0.72 to 0.83 across both splits confirm that the decision chain learns meaningful chemical group information.

\subsection{Additional Case Studies with the Integrated Pipeline}
\label{app:additional_case_studies}
\begin{table}[!htbp]
\centering
\caption{Additional case studies of synthesis plan generation on unseen materials. Cells show \textbf{predicted}\,/\,reported values from \proposed~integrated with the condition model~\citep{huo2022machine}.}
\label{tab:case_studies_appendix}
\small
\setlength{\tabcolsep}{4pt}
\renewcommand{\arraystretch}{0.90}
\definecolor{caseheader}{RGB}{230, 240, 250}  %
\begin{tabular}{@{}clcc@{}}
\toprule
\textbf{\#} & \textbf{Op.} & \textbf{$T$ (P/R, \textdegree C)} & \textbf{$t$ (P/R, h)} \\
\midrule
\rowcolor{caseheader}\multicolumn{4}{@{}l}{\textbf{Case 3:} Mg$_{0.94}$Ni$_{0.06}$ZrNb$_2$O$_8$ \;\; (DOI: 10.1007/s10854-017-7996-x)} \\
\rowcolor{caseheader}\multicolumn{4}{@{}l}{\textit{Pre.:} MgO, Nb$_2$O$_5$, NiO, ZrO$_2$} \\
\rowcolor{caseheader}\multicolumn{4}{@{}l}{\textit{Ops:} mix$\rightarrow$dry$\rightarrow$heat$\rightarrow$mix$\rightarrow$dry$\rightarrow$mix$\rightarrow$shape$\rightarrow$sinter} \\
\cmidrule(l{0pt}r{0pt}){1-4}
0 & heat   & \textbf{1026}\,/\,1100 & \textbf{4.5}\,/\,2.0 \\
1 & sinter & \textbf{1351}\,/\,1320 & \textbf{4.9}\,/\,4.0 \\
\midrule
\rowcolor{caseheader}\multicolumn{4}{@{}l}{\textbf{Case 4:} Ni$_{0.78}$Cu$_{0.20}$Zn$_{0.02}$Fe$_{1.96}$O$_{3.94}$ \;\; (DOI: 10.1016/j.jmmm.2017.12.108)} \\
\rowcolor{caseheader}\multicolumn{4}{@{}l}{\textit{Pre.:} CuO, Fe$_2$O$_3$, NiO, ZnO} \\
\rowcolor{caseheader}\multicolumn{4}{@{}l}{\textit{Ops:} mix$\rightarrow$dry$\rightarrow$heat$\rightarrow$mix$\rightarrow$dry$\rightarrow$mix$\rightarrow$shape$\rightarrow$sinter} \\
\cmidrule(l{0pt}r{0pt}){1-4}
0 & heat   & \textbf{856}\,/\,800 & \textbf{3.1}\,/\,2.0 \\
1 & sinter & \textbf{989}\,/\,900 & \textbf{2.8}\,/\,2.0 \\
\midrule
\rowcolor{caseheader}\multicolumn{4}{@{}l}{\textbf{Case 5:} Pb$_{0.95}$Ba$_{0.05}$Li$_{0.25}$La$_{0.25}$Mo$_{0.5}$O$_3$ \;\; (DOI: 10.1016/S0167-577X(00)00160-9)} \\
\rowcolor{caseheader}\multicolumn{4}{@{}l}{\textit{Pre.:} BaCO$_3$, La$_2$O$_3$, Li$_2$CO$_3$, MoO$_3$, PbO} \\
\rowcolor{caseheader}\multicolumn{4}{@{}l}{\textit{Ops:} mix$\rightarrow$heat$\rightarrow$mix$\rightarrow$heat$\rightarrow$mix$\rightarrow$shape$\rightarrow$sinter} \\
\cmidrule(l{0pt}r{0pt}){1-4}
0 & heat   & \textbf{623}\,/\,550 & \textbf{11.6}\,/\,12.0 \\
1 & heat   & \textbf{677}\,/\,650 & \textbf{11.1}\,/\,18.0 \\
2 & sinter & \textbf{786}\,/\,650 & \textbf{9.5}\,/\,6.0 \\
\bottomrule
\end{tabular}
\end{table}

To further illustrate the practical applicability of the integrated pipeline, we provide three additional case studies in Table~\ref{tab:case_studies_appendix}, complementing the two cases shown in Table~\ref{tab:case_studies}. In each case, \proposed~successfully predicts the precursor set and operation sequence from the target composition, and the integrated condition prediction model~\citep{huo2022machine} supplies the temperature and duration for each heating operation. For reference, the DOIs of the source publications for the two cases shown in the main paper (Table ~\ref{tab:case_studies}) are: Case~1, 10.1007/s10854-017-6960-0; Case~2, 10.1016/j.ssc.2007.09.002.

\subsection{Performance across Multiple Seeds}
\label{app:multiple_seeds}
Table~\ref{tab:multiple_seeds} reports the mean and standard deviation over three seeds on Split~2 for PP, SOP, and MSP. Both backbones obtain similar average performance, with PP Top-10 above 95\% and SOP Top-10 above 32\%. MSP Top-10 reaches $23.48\pm0.74$ for Qwen and $23.91\pm0.90$ for LLaMA, showing consistent performance across the reported runs.

\begin{table}[htbp]
\centering
\captionsetup{position=top,skip=7pt}
\caption{Performance over three seeds on Split~2 (mean $\pm$ standard deviation). Top-$K$ scores are percentages; NED, LCS, and F-1 are similarities.}
\label{tab:multiple_seeds}
\small
\setlength{\tabcolsep}{5pt}
\renewcommand{\arraystretch}{1.15}
\begin{tabular}{lcccc}
\toprule
\multicolumn{5}{c}{\textbf{(a) Precursor Prediction (PP)}} \\
\midrule
\textbf{Model} & Top-1 & Top-3 & Top-5 & Top-10 \\
\midrule
\proposed~(Qwen-2.5-7B) & $72.73\pm0.58$ & $88.90\pm0.20$ & $93.38\pm0.47$ & $95.68\pm0.22$ \\
\proposed~(LLaMA-3.1-8B) & $73.23\pm0.46$ & $89.33\pm0.36$ & $93.59\pm0.19$ & $96.24\pm0.12$ \\
\midrule
\multicolumn{5}{c}{\textbf{(b) Synthesis Operation Prediction (SOP)}} \\
\midrule
\textbf{Model} & Top-10 & NED$\uparrow$ & LCS$\uparrow$ & F-1$\uparrow$ \\
\midrule
\proposed~(Qwen-2.5-7B) & $33.27\pm0.77$ & $0.7489\pm0.0049$ & $0.8283\pm0.0031$ & $0.8537\pm0.0031$ \\
\proposed~(LLaMA-3.1-8B) & $32.84\pm1.14$ & $0.7483\pm0.0046$ & $0.8286\pm0.0039$ & $0.8536\pm0.0030$ \\
\midrule
\multicolumn{5}{c}{\textbf{(c) Material Synthesis Planning (MSP)}} \\
\midrule
\textbf{Model} & Top-1 & Top-3 & Top-5 & Top-10 \\
\midrule
\proposed~(Qwen-2.5-7B) & $6.96\pm0.31$ & $13.43\pm0.35$ & $17.26\pm0.60$ & $23.48\pm0.74$ \\
\proposed~(LLaMA-3.1-8B) & $7.00\pm0.39$ & $12.93\pm0.58$ & $17.04\pm0.43$ & $23.91\pm0.90$ \\
\bottomrule
\end{tabular}
\end{table}

\subsection{Counterfactual Precursor Substitution}
\label{app:counterfactual}
To examine whether PCF makes the decoder use the input precursors, we replace every carbonate precursor with the corresponding cation oxide (e.g., SrCO$_3\rightarrow$SrO) while keeping the target fixed (603 of 1,072 Split~2 test cases). Since carbonates typically require an initial heating step to remove CO$_2$ (Section~\ref{hierarchical precursor types}), this substitution should reduce heating-family operations (heating, sintering, and annealing). Following the probing analysis in Table~\ref{tab:probing}, both models exclude hierarchical precursor types.

\begin{table}[htbp]
\centering
\captionsetup{position=top,skip=7pt}
\caption{Counterfactual carbonate$\rightarrow$oxide substitution on Split~2 (LLaMA-3.1-8B, $n{=}603$). Heat ops: total number of heating-family operations over the 603 cases (original$\rightarrow$substituted). $\downarrow/{=}/\uparrow$: heating count decreased/unchanged/increased among changed outputs.}
\label{tab:counterfactual}
\small
\setlength{\tabcolsep}{5pt}
\renewcommand{\arraystretch}{1.12}
\begin{tabular}{lccccc}
\toprule
\textbf{Model} & \textbf{Heat ops (total)} & \textbf{$\Delta$abs} & \textbf{$\Delta$rel} & \textbf{$n$ changed} & \textbf{$\downarrow/{=}/\uparrow$ among changed (\%)} \\
\midrule
Implicit & 945$\rightarrow$909 & $-36$ & $-3.8\%$ & 140 & 31 / 55 / 14 \\
Explicit (PCF) & 689$\rightarrow$620 & $-69$ & $-10.0\%$ & 128 & 52 / 39 / 9 \\
\bottomrule
\end{tabular}
\end{table}

As shown in Table~\ref{tab:counterfactual}, heating operations barely change under implicit conditioning ($-3.8\%$) but decrease under PCF ($-10.0\%$). Among the changed outputs, most implicit edits leave the heating count unchanged (55\%), whereas most PCF edits reduce it (52\% vs.\ 9\% increase). Thus, PCF makes the generated operations respond to precursor changes in a chemically consistent direction, complementing the probing results in Table~\ref{tab:probing}.

\section{Limitations}
\label{app: Limitations}
\proposed~ successfully solves both PP and SOP and connects them to address the MSP task. However, in the current setup, only the Top-1 PP prediction is used as input, which means that errors in PP can propagate to the SOP stage. We provide a detailed error propagation analysis in Appendix~\ref{app:error_decomp}, which quantifies the contribution of PP errors to MSP failure. A strategy is needed to enable the use of multiple precursor candidates predicted by PP and to incorporate them into the SOP stage so that the model can select better precursor conditions and improve overall performance. In addition, since PP is restricted to precursors above a frequency threshold, its accuracy decreases for rare precursors, and generating precursors unseen during training remains difficult; a larger corpus with broader precursor coverage may alleviate this.
Moreover, while \proposed~produces precursors and synthesis operations from a target material and can be integrated with downstream condition prediction models, it does not directly model processing conditions. Other synthesis attributes such as atmosphere and equipment are also outside the current scope. These aspects remain underexplored largely due to the lack of consistently curated data; once such data become available, incorporating them into the framework will be an important direction for future work.

\section{Broader Impacts}
\label{app: Impacts}
\noindent\textbf{Positive Societal Impacts.} \proposed~provides a structured LLM framework for autonomous material synthesis planning, predicting precursors and synthesis operations for a target material—tasks that are typically challenging and time-consuming for domain experts. By offering an informative starting point for synthesis, our framework can contribute to accelerating materials discovery and reducing the manual effort required in laboratory practice.

\noindent\textbf{Negative Societal Impacts.} The predicted synthesis plans should be reviewed by human experts before experimental execution, as model outputs may contain errors. We do not foresee additional risks beyond standard considerations for AI-assisted scientific tools.

\section{Supplementary Results}
\label{app: Supplementary}
In this section, we provide additional metrics—Top-1, Top-3, and Top-5—supplementing the existing SOP task evaluation.
\begin{table}[hbt]
\centering
\caption{Supplementary Results: Synthesis operation prediction (SOP) performance on two dataset splits.}
\label{tab:main_operation_whole}
\small

\begingroup
\setlength{\aboverulesep}{0pt}
\setlength{\belowrulesep}{0pt}

\renewcommand{\arraystretch}{1.3}

\resizebox{1.0\columnwidth}{!}{%

\begin{tabular}{l cccc c cccc}
\toprule
\textbf{Method} &
\multicolumn{4}{c}{\textbf{Split 1}} &

& \multicolumn{4}{c}{\textbf{Split 2}} \\

\cmidrule(lr){2-5} \cmidrule(lr){7-10}

&
\textbf{Top-1} &
\textbf{Top-3} &
\textbf{Top-5} &
\textbf{Top-10} &
& 
\textbf{Top-1} &
\textbf{Top-3}&
\textbf{Top-5} &
\textbf{Top-10} \\
\midrule

\rowcolor{gray!12}

\multicolumn{5}{l}{\textit{Specialized Models}} & & \multicolumn{4}{l}{} \\
\midrule
Retrieval based \citet{he2023precursor}
&1.46  & 3.75 & 5.94 &10.42  & &1.12  & 3.54 & 5.50 & 11.19 \\
SPENDE \cite{na2023artificial}
& 3.66& N/A &N/A  & N/A &&5.71 & N/A & N/A & N/A \\

\midrule
\rowcolor{gray!12}
\multicolumn{5}{l}{\textit{Large Language Models (API)}} & & \multicolumn{4}{l}{} \\
\midrule
GPT-4o-mini (20-shot)   &1.73  &3.11  &4.2  & 5.85 & &1.68 &3.26  &4.01  &7.09  \\
GPT-4o-mini (40-shot)     &1.83 &2.74 &3.56 & 5.94 & & 2.15 &4.2  & 5.69 & 7.93 \\
GPT-4o (20-shot)   &0.09 & 1.09  & 2.56  &5.03  & &0.47  &1.03 & 1.77 & 3.82 \\
GPT-4o (40-shot)     &0.18  &1.01  &1.92  &4.34  & & 0.28 &1.03  &2.05  &4.94  \\
GPT-5.1  (20-shot)   &0.73 & 1.55 & 1.65& 2.93 & & 0.75 & 1.31 &1.4  &2.33  \\
GPT-5.1  (40-shot)     &0.91  &1.74  &1.74  &2.74  & &0.75  &1.03 & 1.77 &2.33 \\
Claude Haiku 4.5 (20-shot)    &  0.18& 0.73 & 1.56 & 2.56& &0.47 &0.84 &1.21 & 1.68 \\
Claude Haiku 4.5 (40-shot)     & 0.91 &2.38 &3.29  &4.11  & &0.47&1.49 & 2.52  &3.26  \\

\midrule\midrule
\textbf{\proposed~(Qwen-2.5-7B)}
& \textbf{7.86} & \textbf{15.36} & \textbf{19.65} & \textbf{26.51}
& & \textbf{8.77} & \textbf{18.47} & \underline{22.57} & \textbf{32.18} \\
\textbf{\proposed~(LlaMA-3.1-8B)}
& \underline{6.40} & \underline{13.71} & \underline{19.20} & \textbf{26.51}
& & \underline{8.12} & \underline{17.35} & \textbf{22.58} & \underline{31.44} \\
\bottomrule
\end{tabular}
}
\endgroup
\end{table}
\clearpage
\section{Prompt Templates}
\label{app: prompt}
In this section, we present the prompt templates used to fine-tune LLMs for PP~\ref{PP stage} and SOP~\ref{SOP stage}. Additionally, we provide the prompts used for the LLM baseline for the SOP task.

\begin{table}[ht]
\centering
\caption{Prompt for Fine-tuning LLMs for Precursor Prediction (PP).}
\label{tab:precursor_prompt}
\vspace{1ex}

\begin{tcolorbox}[colback=gray!5, colframe=black!50, boxrule=0.4pt, sharp corners, enhanced, width=0.99\textwidth]

\textbf{System Prompt:} You are an expert solid-state chemist planning a material synthesis. \\
\vspace{1ex}

\textbf{Prompt:} Task: \\
Given the target compound \textcolor{blue}{\{target\_formula\}}, first classify its material group, and you must select the precursor needed to synthesize the target. \\

Typically, two or more precursors are required. \\

The possible material groups are: \\
\texttt{['oxide', 'alloy', 'phosphate', 'pyrophosphate', 'halide', 'oxyhalide', 'carbide', 'nitride', 'selenide', 'sulfide', 'other']}.

\vspace{2ex}
Output format: \\
\texttt{[ ['material group'], ['precursor1', 'precursor2', ..., 'precursorN'] ]}

\vspace{2ex}
Return only the Python list of single-quoted strings with no extra text or explanation.

\end{tcolorbox}
\end{table}

\begin{table}[b]
\centering
\caption{Prompt for Fine-tuning LLMs for Synthesis Operation Prediction (SOP).}
\label{tab:synthesis_prompt}
\vspace{1ex}

\begin{tcolorbox}[colback=gray!5, colframe=black!50, boxrule=0.4pt, sharp corners, enhanced, width=0.99\textwidth]

\textbf{System Prompt:} You are an expert solid-state chemist planning a material synthesis. \\
\vspace{1ex}

\textbf{Prompt:}

{}[PRECURSOR\_TYPES] \\
types\_present: \textcolor{blue}{\{precursor types\}} \\
{}[/PRECURSOR\_TYPES]

\vspace{1ex}
Task: \\
Given the target compound \textcolor{blue}{\{target\_formula\}}, first classify its material group,
then based on the selected precursors \textcolor{blue}{\{precursors\}},
propose one plausible synthesis operation sequence using only the following operations: \\
\texttt{['heating', 'sintering', 'annealing', 'mixing', 'shaping', 'quenching', 'drying']}.

\vspace{2ex}
Synthesis context: \textcolor{blue}{\{keyword\}}

\vspace{2ex}
Typically, two or more precursors are required. \\

The possible material groups are: \\
\texttt{['oxide', 'alloy', 'phosphate', 'pyrophosphate', 'halide', 'oxyhalide', 'carbide', 'nitride', 'selenide', 'sulfide', 'other']}.

\vspace{2ex}
Output format: \\
\texttt{[ ['material group'], ['precursor\_type1', ..., 'precursor\_typeK'], ['precursor1', ..., 'precursorN'], ['op1', ..., 'opM'] ]}

\vspace{2ex}
Return only the Python list of single-quoted strings with no extra text or explanation.

\end{tcolorbox}
\end{table}

\begin{table}[ht]
\centering
\caption{Prompt for LLM Baselines in Synthesis Operation Prediction (SOP). }
\label{tab:sop llm prompt}
\vspace{1ex}

\begin{tcolorbox}[colback=gray!5, colframe=black!50, boxrule=0.4pt, sharp corners, enhanced, width=0.99\textwidth]

\textbf{System Prompt:} You are an expert solid-state chemist specializing in synthesis operations. \\
\vspace{1ex}

\textbf{Prompt:} Given: \\
- Target material: \textcolor{blue}{\{target\_formula\}} \\
- Precursors for the target material: \textcolor{blue}{\{precursors\}} \\

Task: \\
Propose a plausible synthesis operation sequence for preparing the target from the given precursors. \\

Instructions: \\
1) Choose operations only from the following allowed set: \texttt{\{"heating", "sintering", "annealing", "mixing", "shaping", "quenching", "drying"\}}. \\
2) "Heating" refers to general heating processes (e.g., calcination, firing) and explicitly excludes Sintering and Annealing. \\
3) Output only the operation names — do not include temperatures, times, atmospheres, parameters, or explanations. \\
4) Format the answer as a Python list of lists of strings \textbf{exactly}. \\
\quad - Each inner list is a single synthesis operation sequence, e.g., \texttt{["mixing", "heating", "sintering"]}. \\
\quad - Return \textbf{exactly \textcolor{blue}{\{num\_combinations\}} sequences}. \\
5) Each sequence must contain \textbf{one or more} operations; repetitions are allowed when chemically reasonable. \\
6) Do not invent new operation names, abbreviations, or comments. Do not output anything other than the list of lists.

\vspace{1ex}
\# Reference for operation semantics (do NOT include this section in the output): \\
\# - Heating: any general heating process (e.g., calcination, firing) except sintering and annealing. \\
\# - Sintering: high-temperature treatment to densify materials. \\
\# - Annealing: controlled heating/cooling for structural ordering or stress relief.

\vspace{2ex}
Synthesis context: \textcolor{blue}{\{keyword\}}

\vspace{2ex}
\textbf{Examples of synthesis operation sequences from target and its precursors:} \\
\textcolor{blue}{\{examples\}}

\vspace{2ex}
Output format (example shape only; return exactly \textcolor{blue}{\{num\_combinations\}} sequences): \\
\texttt{[ \\
\quad ["mixing", "heating", "sintering"], \\
\quad ["mixing", "drying", "heating", "annealing"], \\
\quad ... \\
]}

\vspace{1ex}
Return only the data as a plain text Python list. Do not use Markdown code fences or language tags, and do not print escaped characters like \textbackslash n. Start with [ and end with ]. Use single quotes for strings. No extra text.

\end{tcolorbox}
\end{table}

\clearpage

\newpage
\section*{NeurIPS Paper Checklist}
\begin{enumerate}

\item {\bf Claims}
    \item[] Question: Do the main claims made in the abstract and introduction accurately reflect the paper's contributions and scope?
    \item[] Answer: \answerYes{}{} %
    \item[] Justification: The abstract and introduction clearly state our contributions: a structured LLM-based framework (MSP-LLM) that addresses both PP and SOP within a single MSP framework, with a material group decision chain and precursor constraint factorization. The claims regarding superior performance over baselines on PP, SOP, and the joint MSP task are supported by the experimental results in Section 4.

\item {\bf Limitations}
    \item[] Question: Does the paper discuss the limitations of the work performed by the authors?
    \item[] Answer: \answerYes{} %
    \item[] Justification: We discuss the limitations of our framework in Appendix~\ref{app: Limitations}.

\item {\bf Theory assumptions and proofs}
    \item[] Question: For each theoretical result, does the paper provide the full set of assumptions and a complete (and correct) proof?
    \item[] Answer: \answerNA{} %
    \item[] Justification: N/A

    \item {\bf Experimental result reproducibility}
    \item[] Question: Does the paper fully disclose all the information needed to reproduce the main experimental results of the paper to the extent that it affects the main claims and/or conclusions of the paper (regardless of whether the code and data are provided or not)?
    \item[] Answer: \answerYes{} %
    \item[] Justification: We include all the details for reproducing the experimental results in Section \ref{sec:experiments} and Appendix \ref{app: Implementation}.

\item {\bf Open access to data and code}
    \item[] Question: Does the paper provide open access to the data and code, with sufficient instructions to faithfully reproduce the main experimental results, as described in supplemental material?
    \item[] Answer: \answerYes{} %
    \item[] Justification: We provide a GitHub repository with the code and sufficient instructions for reproducing our results. The source code for \proposed~is available at~\textcolor{magenta}{\url{https://github.com/HeewoongNoh/MSP-LLM}}.

\item {\bf Experimental setting/details}
    \item[] Question: Does the paper specify all the training and test details (e.g., data splits, hyperparameters, how they were chosen, type of optimizer) necessary to understand the results?
    \item[] Answer: \answerYes %
    \item[] Justification: We provide a description of the data splits in Section~\ref{sec:experiments} and an anonymized GitHub repository with the code, sufficient instructions, and experimental details for reproducing our results.

\item {\bf Experiment statistical significance}
    \item[] Question: Does the paper report error bars suitably and correctly defined or other appropriate information about the statistical significance of the experiments?
    \item[] Answer: \answerYes{} %
    \item[] Justification: We report the mean and standard deviation over three seeds for PP, SOP, and MSP on Split~2 with both LLaMA-3.1-8B and Qwen-2.5-7B in Appendix~\ref{app:multiple_seeds} (Table~\ref{tab:multiple_seeds}).

\item {\bf Experiments compute resources}
    \item[] Question: For each experiment, does the paper provide sufficient information on the computer resources (type of compute workers, memory, time of execution) needed to reproduce the experiments?
    \item[] Answer: \answerYes{} %
    \item[] Justification: We provide information on the necessary computational resources in Appendix~\ref{app: Implementation}.
    
\item {\bf Code of ethics}
    \item[] Question: Does the research conducted in the paper conform, in every respect, with the NeurIPS Code of Ethics \url{https://neurips.cc/public/EthicsGuidelines}?
    \item[] Answer: \answerYes{} %
    \item[] Justification: We have reviewed the NeurIPS Code of Ethics and confirm that our research conforms to it in every respect.

\item {\bf Broader impacts}
    \item[] Question: Does the paper discuss both potential positive societal impacts and negative societal impacts of the work performed?
    \item[] Answer: \answerYes{} %
    \item[] Justification: We discuss the potential positive societal impacts and negative impacts of our work in the Appendix \ref{app: Impacts}.
    
\item {\bf Safeguards}
    \item[] Question: Does the paper describe safeguards that have been put in place for responsible release of data or models that have a high risk for misuse (e.g., pre-trained language models, image generators, or scraped datasets)?
    \item[] Answer: \answerNA{} %
    \item[] Justification: Although we use pretrained language models, they are only fine-tuned to output material synthesis planning predictions, so there are no significant risks of misuse.

\item {\bf Licenses for existing assets}
    \item[] Question: Are the creators or original owners of assets (e.g., code, data, models), used in the paper, properly credited and are the license and terms of use explicitly mentioned and properly respected?
    \item[] Answer: \answerYes{} %
    \item[] Justification: All external assets used in this work, including datasets, pretrained models, and baseline implementations, are properly cited and used in accordance with their respective licenses.

\item {\bf New assets}
    \item[] Question: Are new assets introduced in the paper well documented and is the documentation provided alongside the assets?
    \item[] Answer: \answerYes{} %
    \item[] Justification: We release the implementation of MSP-LLM as an anonymized GitHub repository, containing the code, training scripts, hyperparameter configurations, and instructions for reproducing our results.

\item {\bf Crowdsourcing and research with human subjects}
    \item[] Question: For crowdsourcing experiments and research with human subjects, does the paper include the full text of instructions given to participants and screenshots, if applicable, as well as details about compensation (if any)? 
    \item[] Answer: \answerNA %
    \item[] Justification: This paper does not involve any crowdsourcing nor research with human subjects.

\item {\bf Institutional review board (IRB) approvals or equivalent for research with human subjects}
    \item[] Question: Does the paper describe potential risks incurred by study participants, whether such risks were disclosed to the subjects, and whether Institutional Review Board (IRB) approvals (or an equivalent approval/review based on the requirements of your country or institution) were obtained?
    \item[] Answer: \answerNA{} %
    \item[] Justification: This research does not involve human subjects or crowdsourcing.

\item {\bf Declaration of LLM usage}
    \item[] Question: Does the paper describe the usage of LLMs if it is an important, original, or non-standard component of the core methods in this research? Note that if the LLM is used only for writing, editing, or formatting purposes and does \emph{not} impact the core methodology, scientific rigor, or originality of the research, declaration is not required.
    \item[] Answer: \answerNA{} %
    \item[] Justification: LLMs were used only for grammar checking and editing during manuscript preparation, and did not contribute to the core methodology, scientific content, or originality of the research.

\end{enumerate}

\end{document}